\documentclass[sigconf]{acmart}
\AtBeginDocument{%
  }

\usepackage[capitalize]{cleveref}
\crefname{section}{Sec.}{Secs.}
\Crefname{section}{Section}{Sections}
\Crefname{table}{Table}{Tables}
\crefname{table}{Tab.}{Tabs.}

\copyrightyear{2025}
\acmYear{2025}
\setcopyright{acmlicensed}\acmConference[SA Conference Papers '25]{SIGGRAPH Asia
2025 Conference Papers}{December 15--18, 2025}{Hong Kong, China}
\acmBooktitle{SIGGRAPH Asia 2025 Conference Papers (SA Conference Papers '25),
December 15--18, 2025, Hong Kong, China}
\acmDOI{10.1145/3757377.3763935}
\acmISBN{979-8-4007-2137-3/2025/12}

\begin{document}

\title[MV-Performer]{\textit{MV-Performer}: Taming Video Diffusion Model for Faithful and Synchronized Multi-view Performer Synthesis}

\author{Yihao Zhi}
\orcid{0009-0007-1183-5459}
\affiliation{%
 \institution{SSE, CUHKSZ}
 \city{Shenzhen}
 \country{China}}
\email{yihaozhi1@link.cuhk.edu.cn}

\author{Chenghong Li}
\orcid{0009-0004-0604-7421}
\affiliation{%
 \institution{FNii-Shenzhen and SSE, CUHKSZ}
 \city{Shenzhen}
 \country{China}}
\email{chenghongli@link.cuhk.edu.cn}

\author{Hongjie Liao}
\orcid{0009-0005-1992-7089}
\affiliation{%
 \institution{SSE, CUHKSZ}
 \city{Shenzhen}
 \country{China}}
\email{hongjieliao@link.cuhk.edu.cn}

\author{Xihe Yang}
\orcid{0009-0009-0066-9365}
\affiliation{%
 \institution{SSE, CUHKSZ}
 \city{Shenzhen}
 \country{China}}
\email{xiheyang1@link.cuhk.edu.cn}

\author{Zhengwentai Sun}
\orcid{0000-0002-3884-137X}
\affiliation{%
 \institution{SSE, CUHKSZ}
 \city{Shenzhen}
 \country{China}}
\email{zhengwentaisun@link.cuhk.edu.cn}

\author{Jiahao Chang}
\orcid{0009-0009-6877-1649}
\affiliation{%
 \institution{SSE, CUHKSZ}
 \city{Shenzhen}
 \country{China}}
\email{jiahaochang@link.cuhk.edu.cn}

\author{Xiaodong Cun}
\orcid{0000-0003-3607-2236}
\affiliation{%
 \institution{Great Bay University}
 \city{Dongguan}
 \country{China}}
\email{cun@gbu.edu.cn}

\author{Wensen Feng}
\orcid{0000-0002-7315-337X}
\affiliation{%
 \institution{School of Artificial Intelligence, Shenzhen University}
 \city{Shenzhen}
 \country{China}}
\email{sanmumuren@126.com}

\author{Xiaoguang Han}
\orcid{0000-0003-0162-3296}
\affiliation{%
 \institution{SSE, CUHKSZ and FNii-Shenzhen and Guangdong Provincial Key Laboratory of Future Networks of Intelligence}
 \city{Shenzhen}
 \country{China}}
\email{hanxiaoguang@cuhk.edu.cn}

\renewcommand{\shortauthors}{Zhi, et al.}

\authornote{Corresponding author: Xiaoguang Han (hanxiaoguang@cuhk.edu.cn).}

\newcommand{\PIPENAME}{MV-Performer}
%
\begin{abstract}
Recent breakthroughs in video generation, powered by large-scale datasets and diffusion techniques, have shown that video diffusion models can function as implicit 4D novel view synthesizers. Nevertheless, current methods primarily concentrate on redirecting camera trajectory within the front view while struggling to generate 360-degree viewpoint changes. In this paper, we focus on human-centric subdomain and present \PIPENAME, an innovative framework for creating synchronized novel view videos from monocular full-body captures. To achieve a 360-degree synthesis, we extensively leverage the MVHumanNet dataset and incorporate an informative condition signal. Specifically, we use the camera-dependent normal maps rendered from oriented partial point clouds, which effectively alleviate the ambiguity between seen and unseen observations. To maintain synchronization in the generated videos, we propose a multi-view human-centric video diffusion model that fuses information from the reference video, partial rendering, and different viewpoints. Additionally, we provide a robust inference procedure for in-the-wild video cases, which greatly mitigates the artifacts induced by imperfect monocular depth estimation. Extensive experiments on three datasets demonstrate our \PIPENAME ’s state-of-the-art effectiveness and robustness, setting a strong model for human-centric 4D novel view synthesis. Code is available at \href{https://github.com/zyhbili/MV-Performer}{https://github.com/zyhbili/MV-Performer}.
\end{abstract}



\begin{teaserfigure}
\centering
  \includegraphics[trim=0cm 0cm 0cm 0cm, clip=true,width=\linewidth]{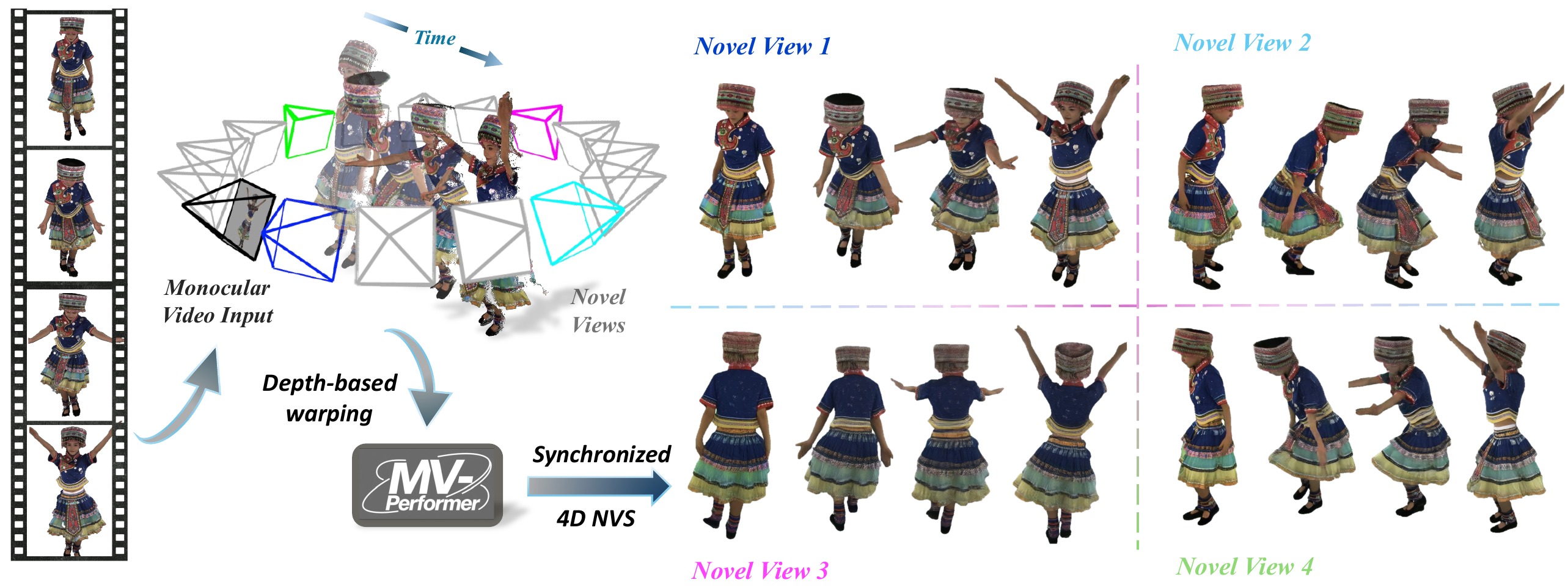}
  \vspace{-6mm}
  \caption{We propose \textit{\PIPENAME{}} that aims to generate 4D human novel view synthesis from monocular video input. Our method adopts the powerful video diffusion model with the depth-based warping paradigm, enabling 360-degree synchronized multi-view video generation. \PIPENAME~demonstrates strong capabilities in maintaining both view and temporal consistency for 4D human novel view synthesis.}
  \label{fig:teaser}
\end{teaserfigure}

\maketitle

\section{Introduction}
\label{sec:intro}
Novel view synthesis is a longstanding task in 3D vision and computer graphics, with extensive applications in media content creation, augmented and virtual reality, movie production, etc. Early methods~\cite{avidan1997novel,chaurasia2011silhouette,chen2023view,levoy2023light} attempt to solve it with techniques including multi-view stereo~\cite{seitz2006comparison,furukawa2015multi} and image warping~\cite{glasbey1998review}, which explicitly model the stereo, color of each target pixel. With the rise of neural representations and corresponding differentiable rendering techniques~\cite{tewari2020state,thies2019deferred,park2019deepsdf,nerf,Chen2022ECCV,shen2021dmtet,jiang2020sdfdiff,kerbl20233d,Huang2DGS2024}, high-fidelity novel view synthesis can be obtained through reconstruction from posed visual observations. However, fine reconstructions often require high capture coverage and density.

Beyond static scenes, a comprehensive 4D human synthesis \cite{orts2016holoportation,hilsmann2020going,xu20244k4d,li2024animatable}, viewable from all angles, is more crucial for enhancing immersive experiences. However, 4D human reconstruction presents unique challenges because of its ill-posedness. For example, a static scene can be thoroughly documented over time using a smartphone; however, when it comes to a person in motion, we are limited to capturing only a partial snapshot at one moment with the same device. Therefore, 4D human novel view synthesis generally demands a synchronized and calibrated multi-view camera system \cite{2023dnarendering,xiong2024mvhumannet, li2025mvhumannet++}, which is both costly and sophisticated. Motivated by recent advancements in techniques \cite{wan2025} and datasets \cite{xiong2024mvhumannet, li2025mvhumannet++}, we believe it is the opportune moment to make a breakthrough: realizing a 360-degree human-centric dynamic novel view synthesis using only monocular inputs.

Diffusion Probabilistic Models~\cite{sohl2015deep,song2019generative,ho2020denoising} have witnessed huge success in recent years, particularly for image and video generation tasks. Certain diffusion-based models possess the capability to infer and generate the shape and appearance of an object's multiple views from a single frontal image, maintaining high spatial consistency~\cite{watson2022novel,liu2023zero,liu2023one,shi2023mvdream,liu2024one,shi2023zero123++,liu2023syncdreamer,wang2023imagedream,voleti2024sv3d,Kant2024Pippo}. Building upon these multi-view diffusion models, 4D generation is attainable by additionally enforcing the temporal consistency~\cite{jiang2023consistent4d,ling2024align,ren2023dreamgaussian4d,zeng2024stag4d,bahmani20244d,wu2024sc4d,huang2025mvtokenflow} through 4D representations \cite{fridovich2023k,wu20244d}. Although similar strategies can be directly applied to 4D human scenarios \cite{Disco4D}, their training processes are still expensive, and they remain inadequate for handling large motions and preserving temporal details due to limitations inherent in their foundation models.

\begin{figure}[t]
    \centering
\includegraphics[width=\columnwidth]{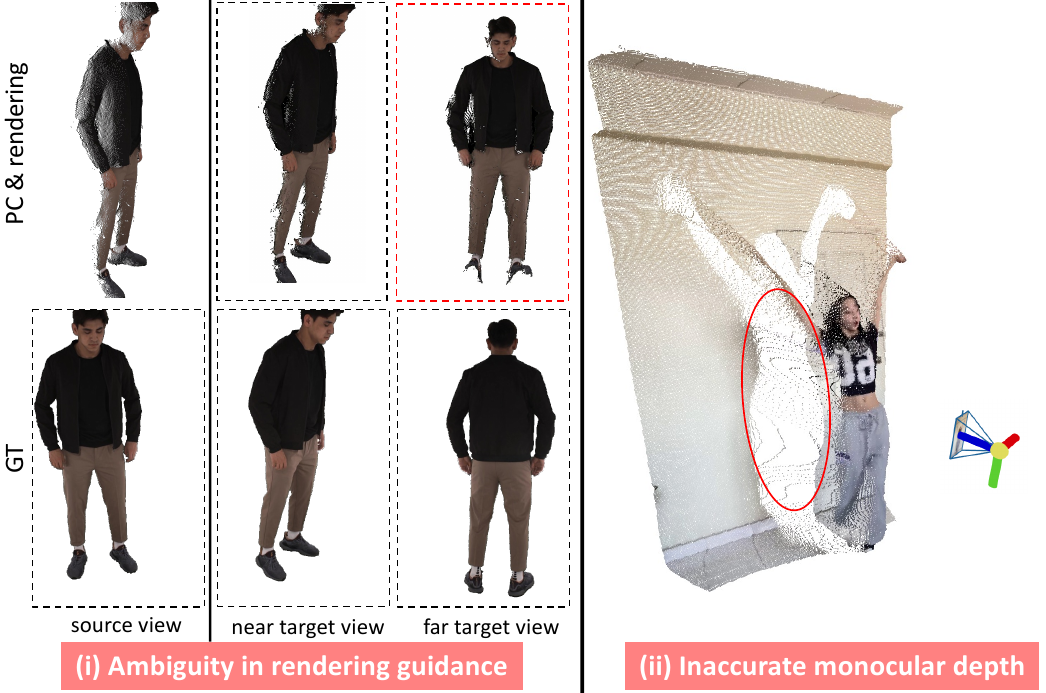}
\vspace{-6mm}
    \caption{(i) The depth warping condition at the rear viewpoints presents ambiguity for the model. (ii) Inaccurate monocular depth produce floater-like rendering when there is a significant change in viewpoint.}
    \label{fig:challenges}
\end{figure}

Recent rapid evolution of video diffusion model \cite{blattmann2023stable,blattmann2023align,chen2024videocrafter2,chen2023videocrafter1,xing2023dynamicrafter,he2022lvdm,lin2024open,rombach2022high,hong2022cogvideo,yang2024cogvideox,wan2025} demonstrates its potential to function as a shader \cite{gu2025das} and enable camera-controllable video generation \cite{wang2024motionctrl,he2024cameractrl,wu2024cat4d}. It is possible to directly infer novel view video content through iteratively sampling and denoising, obviating the need for scene-specific training. Some works~\cite{vanhoorick2024gcd,bai2024syncammaster,bai2025recammaster,jiang2024animate3d} redirect the camera trajectory via the injection of camera pose embeddings. However, these models generally converge at a relatively slow pace. Moreover, such an implicit condition typically demands a dense array of viewpoints in the training set to guarantee generalizability across arbitrary perspectives. Another line of works \cite{yu2024viewcrafter,ren2025gen3c,yu2025trajectorycrafter,bian2025gs,liu2025free4d,xiang20233d} explicitly employ depth geometric priors. They achieve 4D novel view synthesis by first applying depth-based warping and then employing video inpainting. Despite these successes, these works struggle to synthesize at very large viewpoint changes and faithfully preserve multi-view attributes. Apart from the limitations of training data, the reasons are still twofold (\cref{fig:challenges}): $(i)$ insufficient 3D cues from monocular inputs are provided to the network. $(ii)$ image warping floater at large viewpoints change would be intolerable due to inaccurate monocular depth estimation.

In this paper, we focus on human-centric scenarios and present \PIPENAME, a simple yet effective framework that transforms an input monocular video into multi-view synchronized videos. In particular, we extend the pre-trained WAN2.1 \cite{wan2025} to a multi-view video diffusion model that learns the joint distribution of multi-view human-centric videos. 
To address the aforementioned issues, we devise a network tailored to the data characteristics of MVHumanNet~\cite{xiong2024mvhumannet}. We contend that using implicit camera embeddings is unsuitable for MVHumanNet \cite{xiong2024mvhumannet} due to the limited camera views. 
To enable a 360-degree novel view synthesis with the explicit depth-based warping paradigm, we excavate additional condition information from the monocular depth. Specifically, we render the camera-dependent normal map from oriented point clouds, which aid the model in distinguishing between observed and unobserved areas. To ensure synchronization within different views and faithfulness toward the reference view, our multi-view video diffusion model adopts the multi-view attention and reference attention mechanisms, which efficiently fuse the information from the reference video, partial rendering, and different viewpoints. Additionally, we provide a robust inference procedure by integrating several state-of-the-art estimation methods \cite{piccinelli2025unidepthv2,li2024_megasam,khirodkar2024sapiens}, which significantly mitigate the artifacts induced by imperfect monocular depth estimation and provide better guidance to video generation.

 Extensive experiments on MVHumanNet \cite{xiong2024mvhumannet}, DNA-Rendering \cite{2023dnarendering}, and collected in-the-wild datasets demonstrate the superior effectiveness and robustness of our proposed \PIPENAME. In summary, our contributions are as follows:

\begin{itemize}
    \item We develop the first generative framework for converting human-centric monocular video to dense multi-view videos, leveraging a cutting-edge video diffusion model and the MVHumanNet dataset.

    \item We propose a multi-view video diffusion model that learns the joint distribution of multi-view human-centric videos, guided by the normal map rendered from oriented partial point clouds. We show that the depth-based warping paradigm could also enable human appearance and motion synthesis under large viewpoint changes, harnessing the inherent power of the video diffusion model.

    \item To ensure the generalizability of our framework, we provide a robust inference procedure, which greatly mitigates the artifacts induced by imperfect monocular depth estimation.

\end{itemize}

\section{Related work}
\label{sec:related}

\subsection{Reconstruction-based 4D Human Modeling}

4D novel view synthesis presents significant challenges, which are typically achieved by first reconstructing the dynamic scenes. Numerous highly efficient and expressive 4D representations \cite{cao2023hexplane, huang2024sc, lin2024gaussian, shao2023tensor4d,duan20244d,yang2023gs4d,li2024spacetime,xu2024longvolcap,wang2025freetimegs} are introduced to improve reconstruction performance.
Recently, high-fidelity 4D human reconstruction has been widely investigated to achieve photorealistic digital avatar creation. Multi-view approaches, designed for studio environments with calibrated sensors, leverage diverse scene representations—such as volumetric occupancy fields~\cite{huang2018deep}, point clouds~\cite{wu2020multi}, and depth fusion~\cite{yu2021function4d}—to capture clothed human performances. The success of neural radiance fields (NeRF)~\cite{mildenhall2020nerf} further advanced this domain,  follow-up works~\cite{peng2021neural, peng2021animatable, liu2021neural,li2022tava, wang2022arah, zhi2022dual,zheng2022structured,zhao2022human, zheng2023avatarrex, li2023posevocab} utilize neural rendering techniques to learn a plausible implicit canonical geometry~\cite{pumarola2021d} of clothed humans. while recent work explores 3D Gaussian splatting~\cite{kerbl20233d} for efficient photo-realistic human rendering~\cite{pang2024ash, li2024animatable, jiang2024hifi4g, chen2024meshavatar, jiang2024robust,Qian_2024_CVPR, chen2025taoavatar}. However, these methods rely on specialized hardware, restricting their applicability. 
In contrast, monocular reconstruction tackles the ill-posed challenge of inferring 3D geometry from single-view inputs \cite{zhao2025surfel,WangCVPR2024,kocabas2024hugs}. NeRF-based works~\cite{jiang2022selfrecon, jiang2022neuman, weng2022humannerf, guo2023vid2avatar, jiang2023instantavatar} adopted neural deformation fields to model dynamic humans from monocular videos. Inspired by these methods, recent advances~\cite{hu2024gauhuman, qian20233dgsavatar, zhi2025strugauavatar, wen2024gomavatar} optimize 3DGS primitives anchored to explicit~\cite{loper2015smpl, pavlakos2019smplx} or implicit  templates~\cite{yariv2021volume, wang2021neus, shen2021deep}, achieving articulated avatars with enhanced detail. However, such optimization-based frameworks typically require extensive optimization time to achieve satisfactory performance.

\subsection{Generalizable 4D Human Novel View Synthesis}

Neural rendering technologies~\cite{mildenhall2020nerf, tewari2020state} have demonstrated strong capabilities in generating high-fidelity renderings across multiple views. However, these methods are typically optimized for a single scene and require densely sampled input views for training. For general scenes, some representative works \cite{xu2022point, mvsnerf, chen2024mvsplat} follow the multi-view stereo fashion and propose generic deep neural networks to directly regress neural parameters. To extend their applicability to new human performers and handle sparse-view inputs, later works~\cite{kwon2021neural, chen2022geometry, zhao2022humannerf, mihajlovic2022keypointnerf, hu2023sherf} use 3D human prior to anchor the pixel-aligned features accurately on the human template. Although these techniques achieve good results, their rendering speed is slow due to the heavy computations in volume rendering. Recent methods~\cite{zheng2024gps, kwon2024generalizable, zhuang2024idol, hu2024eva} utilize GPU-accelerated 3DGS rasterization~\cite{kerbl20233d} to achieve both high-speed and photorealistic human rendering from sparse observations. Nevertheless, these methods can only generate promising results for observed viewpoints and still struggle to synthesize fine details in unseen regions.

\subsection{4D View Extrapolation via Video Diffusion Models}
Diffusion models~\cite{ho2020denoising, rombach2022high, song2020denoising} have demonstrated remarkable promise in generating novel views from posed sparse view videos \cite{jin2025diffuman4d} or even from a monocular video. One line of works~\cite{he2024cameractrl, bai2025recammaster} encoding camera pose parameters into the video diffusion models for controlling the viewpoint of the output video. In another line, GEN3C \cite{ren2025gen3c}, TrajactoryCrafter \cite{yu2025trajectorycrafter}, and others \cite{hu2025ex4dextremeviewpoint4d,bian2025gs} converge on the concept of employing depth-based warping information as prior conditions. However, these models cannot effectively generate synchronized multi-view videos consistent with each other. Recent studies have extended beyond single-camera scenarios, focusing on multi-view video generation. SV4D~\cite{xie2024sv4d} and CAT4D~\cite{wu2024cat4d} combine 3D shape and motion information from multi-view video diffusion to optimize implicit 4D representations. SynCamMaster~\cite{bai2024syncammaster} introduces a multi-view synchronization module to synthesize open-world multi-view videos from a single text prompt and desired viewpoints. For multi-view human video generation, Human4DiT~\cite{shao2024human4dit} introduces a 4D diffusion transformer that disentangles image, viewpoint, and temporal learning. 
GAS~\cite{lu2025gas} employs video diffusion models to enhance novel-view and pose synthesis results from Human NeRF reconstruction. However, these models primarily focus on pose-conditioned human animation from single-image inputs rather than 4D novel view synthesis from monocular videos, and some codes are not publicly available.
\section{Preliminary}
\subsection{Flow Matching}
Flow matching models~\cite{lipman2022flow, esser2024scaling} synthesize data by continuously transforming a simple noise distribution into a complex target distribution through an ordinary differential equation (ODE). At time $t \in [0, 1]$, the model evolves a sample $\mathbf{x}(t) \in \mathbb{R}^d$ and may optionally condition on auxiliary information $c$ (e.g., text embeddings or reference images).

Given a pair of points $\mathbf{x}_0 \sim \mathcal{N}(0, \mathbf{I})$ and $\mathbf{x}_1 \sim p_{\text{data}}$, a linear interpolation is defiend as follows:
\begin{equation}
\mathbf{x}_t = (1 - t)\mathbf{x}_0 + t\mathbf{x}_1.
\end{equation}
The model learns a velocity field $v_\theta: \mathbb{R}^d \times \mathcal{C} \times [0,1] \rightarrow \mathbb{R}^d$ that predicts the constant displacement vector $\mathbf{v}_t = \mathbf{x}_1 - \mathbf{x}_0$. The training objective minimizes the expected squared error:
\begin{equation}
\mathcal{L} = \mathbb{E}_{\mathbf{x}_0, \mathbf{x}_1, c, t} \left\| v_\theta(\mathbf{x}_t, c, t) - (\mathbf{x}_1 - \mathbf{x}_0) \right\|^2.
\end{equation}
Once trained, generation is performed by solving the ODE:
\begin{equation}
\frac{d\mathbf{x}(t)}{dt} = v_\theta(\mathbf{x}(t), c, t), \quad \mathbf{x}(0) \sim \mathcal{N}(0, \mathbf{I}),
\end{equation}
from $t=0$ to $t=1$, yielding $\mathbf{x}(1)$ as the final output. Compared to DDPM~\cite{ho2020denoising}, this formulation allows efficient sample generation with substantially fewer integration steps.

\subsection{WAN 2.1}
To ensure temporal consistency in the generated results, we adopt WAN 2.1 \cite{wan2025wan} as our backbone, which is based on flow matching. A key component of this framework is a 3D VAE that jointly encodes video frames into a temporally-aware latent space, enforcing causality while reducing memory consumption. Given a video with $f$ frames and a resolution of $(H, W)$, the 3D VAE compresses it into a latent representation with shape $[1 + f/4, H/8, W/8,  C]$, where $C$ denotes the number of channels. In this latent space, a Diffusion Transformer (DiT) model is employed for video generation, leveraging both temporal structure and a compact representation. To reduce memory consumption, we adopt the 1.3B-parameter version of DiT for training our \PIPENAME~at a resolution of 480px.

\begin{figure*}[t]
\centering
\includegraphics[width=1.0\linewidth]{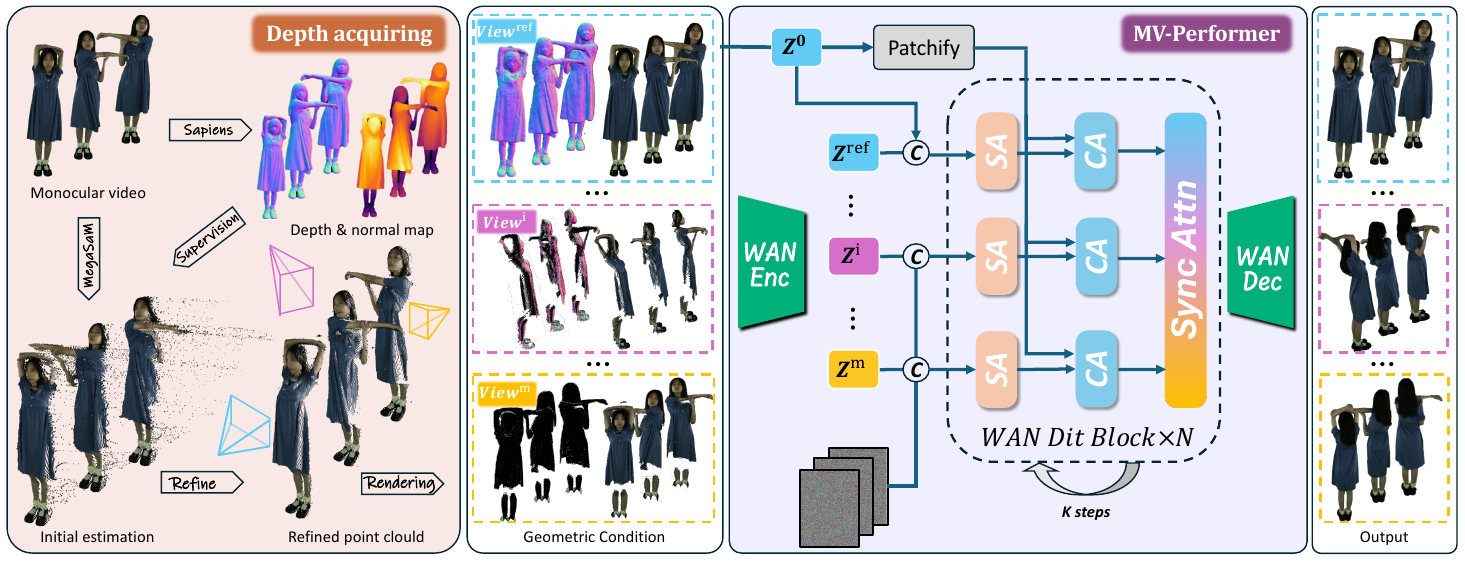}
\caption{The overview of our \PIPENAME.  ``SA''and ``CA'' are abbreviations for self-attention and cross-attention, respectively. We first estimate the depth and normal from Sapiens~\cite{khirodkar2024sapiens} and then use these estimates to refine the noisy point cloud output from MegaSaM~\cite{li2024_megasam}. Next, we render the refined point cloud with corresponding colors to novel views as geometric conditions. Finally, we feed them into~\PIPENAME~to synthesize a 4D human video from novel viewpoints.
}
\label{fig:pipeline}
\end{figure*}

\section{Method}
Given a reference frontal full-body monocular video $V^{ref}$, comprising $f$ frames, our goal is to synthesize $m$ synchronized novel view human videos $\left\{V^1, V^2,...,V^m \right\}$. These videos should accurately maintain consistency across different views. We tackle this problem by taming the power of MVHumanNet \cite{xiong2024mvhumannet} and pre-trained Wan2.1-T2V-1.3B \cite{wan2025}. In this section, we first introduce our 
synchronized multi-view video diffusion model (\cref{sec:mv-vdm}). Then, we illustrate our camera-dependent normal map designed to handle large viewpoint changes (\cref{sec:normal_condition}). Finally, we present the inference procedures for in-the-wild scenarios(\cref{sec:inference}).

\subsection{Multi-View Video Diffusion Model with Depth-based Geometric Condition}
\label{sec:mv-vdm}
The overview of our pipeline is shown in \cref{fig:pipeline}. One primary focus of our design is selecting an appropriate condition according to the dataset characteristics.

\noindent \textbf{Depth-based warping.} As mentioned in \cref{sec:intro}, the open-source multi-view datasets typically comprise 32 to 60 camera views, with the cameras fixed on capture cages. This setup results in a limited training view distribution. Therefore, instead of utilizing Plücker ray as the camera embedding \cite{bai2024syncammaster,bai2025recammaster,hecameractrl}, we incorporate explicit 3D geometric priors for the precise control of camera viewpoint changes, following the depth-based warping paradigm used in \cite{yu2024viewcrafter,yu2025trajectorycrafter,bian2025gs,ren2025gen3c}. To construct the training pairs, we perform RGBD-warping with known camera parameters $\left\{Cam^{ref}, Cam^1,..., Cam^m \right\}$. Specifically, given a frontal RGB image with its metric depth $D$, and corresponding camera parameter $Cam^{ref}$ consisting of intrinsics $K$ and extrinsics $R$, we first unproject the 2D pixels $u$ into the colored partial point cloud $X_{color}$ in the world coordinate:

\begin{equation}
    X(u) = R^{-1}D(u)K^{-1}u
\end{equation}

Subsequently, given new viewpoints $\left\{Cam^1, Cam^2,...,Cam^m \right\}$, we render the per-frame colored point cloud into partial rendering $\mathcal{R}(X_{color}, Cam^i)$ for viewpoint $i$. In this way, we produce the partial rendering geometric cues for $m$ target views $\left\{P^1, P^2,..., P^{m} \right\}$. We feed the partial rendering $\left\{P^1, P^2,..., P^{m} \right\}$ and normal $\left\{N^1, N^2,..., N^m \right\}$ (see \cref{sec:normal_condition}) geometric condition to the 3D-VAE of Wan2.1 separately and concatenate their output along the channel dimension, resulting in latent features $Z^{i}_{cond}$. We further concatenate them with the input noise latents $Z_{noise}$ along the channel dimension. 


For network finetuning, we adhere to the principle of simplicity. To achieve faithful and synchronized generation, we specifically modify the pre-trained Wan2.1-T2V-1.3B model \cite{wan2025} by incorporating two primary components in each DiT block:

\noindent \textbf{Ref Attention.} The partial rendering explicitly represents the camera transformation and effectively provides the denoising network with known observations. However, some information will inevitably be lost due to occlusion. Inspired by \cite{yu2025trajectorycrafter}, we implement cross-attention mechanisms between $Z_{in}$ and reference latents $Z^{ref}$, where $Z_{in}$ denotes the hidden latents in each Dit block. We use $Z_{in}$ as queries and the $Z^{ref}$ as keys and values. The reference latents $Z_{ref}$ are derived from $V^{ref}$ via the VAE encoder in conjunction with a reference patch embedder. 
\begin{equation}
    Z_{out} = Z_{in}+ proj(cross\_attn(Z_{in}, Z_{ref}))   
\end{equation}

The aggregated features are projected back to the original dimension with a zero-initialized linear layer and residual connection. Unlike \citet{yu2025trajectorycrafter}, which incorporates additional attention layers, we reuse the textual cross attention layer for simplicity.  

\noindent \textbf{Sync Attention.} Despite the consistent underlying 3D geometry $P_{color}$, challenges persist in maintaining consistency across various camera viewpoints. This issue is particularly pronounced when considering views from the rear. To aggregate information from the hidden latents $Z_{in} = concat(Z^{ref}_{in}, Z^1_{in},...,Z^m_{in})$ across different viewpoints, where $m$ is the target number of views. We employ a frame-level spatial self-attention mechanism that functions as synchronized attention: 

\begin{equation}
    Z_{out} = Z_{in}+ proj(self\_attn(Z^{ref}_{in}, Z^1_{in},...,Z^m_{in}))   
\end{equation}

The synchronized attention mechanism effectively aggregates per-frame information from multiple views and integrates it into the video diffusion model. Unlike \citet{bai2024syncammaster}, we do not incorporate camera pose embedding into our model.

\subsection{Camera-dependent Normal Map Condition}
\label{sec:normal_condition}
The previous warping-based method can only handle small viewpoint changes \cite{xiang20233d, yu2025trajectorycrafter}. We attribute this limitation to the ambiguity between front and back perspectives under larger viewpoint changes. To address this issue, we propose to leverage camera-dependent normal map condition to facilitate 360-degree synthesis. As illustrated in~\cref{fig:pipeline}, we adopt a view-dependent rendering strategy to provide an intuitive representation of surface orientation.  Specifically, given the point cloud normal vector 
$\vec{n}$  and the camera viewing direction $\vec{d}$, both defined in the world coordinate system (with $\vec{d}$ derived from the camera's rotation matrix), we compute the dot product $o = \vec{n} \cdot \vec{d}$ for each point.
The value of $o$ indicates the surface orientation: $o>0$ implies the surface is facing the camera, while $o<0$ denotes it is facing away. For visualization, we map the normal vectors from the $[-1,1]$ range to the RGB color space $[0,1]$, and assign black to surfaces where $o<0$, effectively masking back-facing areas. This strategy not only highlights the geometric structure of the point cloud but also conveys precise orientation cues, which are critical for accurate multi-view synthesis. We denote the camera-dependent normal map rendering videos as $\left\{N^1, N^2,...,N^m \right\}$ for $m$ target views.

\subsection{Inference with Refined Monocular Depth}
\label{sec:inference}
For in-the-wild inference, we need to perform the depth-based warping to get the partial rendering of the novel view, necessitating a metric depth estimation method. However, existing approaches \cite{piccinelli2025unidepthv2} continue to face challenges in producing high-fidelity depth outputs. Specifically, the depth drift toward the background significantly degrades the generation quality for large viewpoint changes, mainly due to the domain gap.
To tackle this issue, we propose a depth refinement process by integrating several state-of-the-art estimation methods. Specifically, given a monocular video input $V^{ref} = \left\{I_0, I_1, ..., I_f \right\}$ comprising $f$ frames, we first estimate the per-frame unified metric depth $\hat{D}_i$ and camera parameters using MegaSaM \cite{li2024_megasam}, and the high-quality relative depth $\tilde{D}_i$ normal map $\tilde{N}$ using Sapiens \cite{khirodkar2024sapiens}. Then, we align the relative depth $\tilde{D}_i$ to the coarse metric depth $\hat{D}_i$:

\begin{equation}
    \mathop{\arg\min}_{\alpha,\beta}= || (\alpha \cdot \tilde{D}_i+ \beta) - \hat{D}_i||_2
\end{equation}
This can be effectively solved for scale and shift with a least-squares criterion which has a closed-form solution \cite{yu2022monosdf}. Finally, we further optimize the aligned depth using normal map $\tilde{N}$ \cite{Huang2DGS2024, bini2022cao}. 
\section{Experiments}
\subsection{Datasets}
To access quantitative metrics, we conduct experiments on two extensively used multi-view human modeling datasets, MVHumanNet~\cite{xiong2024mvhumannet} and DNA-Rendering~\cite{2023dnarendering}. \textbf{We only use the training part of MVHumanNet~\cite{xiong2024mvhumannet} as training set.} Additionally, we collect 5 monocular videos from Bilibili and TikTok to demonstrate generalizability.

\textbf{MVHumanNet.} 
MVHumanNet~\cite{xiong2024mvhumannet} is a multi-view video dataset with over 9000 identities in everyday clothing. MVHumanNet++~\cite{li2025mvhumannet++}, an expanded version of MVHumanNet, offers additional depth, normal estimations, and more robust mask segmentation and SMPLX fitting. We utilized 16-view videos from a training set comprising 5,400 subjects for our training process. For evaluation purposes, we selected a test set consisting of 10 subjects. In this test set, we employed even-numbered views to conduct the assessment.

\textbf{DNA-Rendering.} 
DNA-Rendering~\cite{2023dnarendering}, another multi-view video dataset, features some professional actors and complicated clothing. In alignment with the MVHumanNet evaluation setup, we sampled 10 subjects from the 8 camera views subset. This dataset is utilized for evaluation purposes.

\subsection{Baselines}
To the best of our knowledge, we are among the first to concentrate on the subdomain of 360-degree, human-centric 4D novel view synthesis from monocular input. As a result, there are limited established methods available for direct benchmarking.

We mainly compare \PIPENAME{} with three baselines: TrajectoryCrafter \cite{yu2025trajectorycrafter},  ReCamMaster \cite{bai2025recammaster}, and Champ \cite{zhu2024champ},  
where the first two methods are the state-of-the-art, open-sourced camera-controlled video diffusion models, and the last one is the human image animation method. We finetuned ReCamMaster \cite{bai2025recammaster} on MVHumanNet \cite{xiong2024mvhumannet} for 20 epochs to make a fairer comparison.

We do not compare to Human4Dit \cite{shao2024human4dit}, and Disco4D \cite{Disco4D} because they primarily focus on animation rather than 4D novel view synthesis. Moreover, they have not provided open-source code, and we face difficulties affording the training costs for reproducing Human4Dit \cite{shao2024human4dit}.

\subsection{Evaluation Metrics}
To quantitatively evaluate the quality of generated multi-view videos, we report five standard metrics that jointly assess spatial fidelity, perceptual realism, and temporal consistency: PSNR \cite{psnr}, SSIM \cite{ssim}, LPIPS \cite{lpips}, FID \cite{heusel2017gans}, and FVD \cite{unterthiner2018towards}.

PSNR and SSIM measure low-level pixel and structural accuracy with respect to the ground truth views. LPIPS evaluates perceptual similarity using deep features and better reflects human visual judgment.
To assess cross-view coherence and realism at the sequence level, we adopt FID for image distribution alignment, and FVD to measure temporal consistency and holistic video quality using pretrained spatio-temporal features. We compute FVD within the paired ground-truth and generated video sets.

\subsection{Implementation Details}
As noted by \citet{bai2024syncammaster}, we also encounter challenges in directly optimizing our full pipeline. To address this, we implement a progressive training strategy. Our formulation allows for a natural decoupling of the pipeline into two distinct stages: first, video inpainting, followed by synchronization. In the initial stage, we refrain from incorporating the synchronization module and train all other parameters for 5 epochs. In the subsequent stage, our focus shifts to synchronization; thus, we freeze all other modules and exclusively train the synchronization module for an additional 5 epochs. Throughout both training phases, we utilize the AdamW \cite{loshchilov2017decoupled} optimizer set the learning rate at $1 \times 10^{-4}$ and gradually decrease it to $2 \times 10^{-5}$. All experiments are conducted with an effective batch size of $6 \times 12$ on 6 NVIDIA A100. We perform $K=50$ steps sampling for all experiments. Our full pipeline can simultaneously generate around 10 videos with 49 frames on a custom-level GPU with 24G memory like RTX3090.

\begin{table}[t]
\centering
\renewcommand{\arraystretch}{1.2}
\resizebox{\columnwidth}{!}{
\begin{tabular}{l|ccccc}
\toprule
Methods & PSNR ↑ & SSIM ↑ & LPIPS ↓ & FID ↓ & FVD ↓ \\
\midrule
\multicolumn{6}{c}{\textbf{MVHumanNet~\cite{xiong2024mvhumannet}}} \\
\midrule
Champ & 11.23 & 0.813 & 0.328 & 55.92 & 5.54 \\
ReCamMaster & 6.97 & 0.600 & 0.620 & 154.03 & 10.78 \\
ReCamMaster* & 11.62 & 0.817 & 0.287 & 26.44 & 2.17 \\
TrajectoryCrafter & 4.18 & 0.493 & 0.722 & 154.00 & 17.25 \\
\textbf{Ours} &
\textbf{24.35} & \textbf{0.926} & \textbf{0.066} & \textbf{24.47} & \textbf{0.12} \\
\midrule
\multicolumn{6}{c}{\textbf{DNA-Rendering~\cite{2023dnarendering}}} \\
\midrule
Champ & 9.08 & 0.750 & 0.399 & 58.59 & 4.73 \\
ReCamMaster & 6.46 & 0.595 & 0.602 & 138.25 & 7.80 \\
ReCamMaster* & 10.02 & 0.769 & 0.342 & 36.78 & 4.28 \\

TrajectoryCrafter & 4.72 & 0.498 & 0.758 & 154.66 & 15.52 \\
\textbf{Ours} &
\textbf{15.63} & \textbf{0.861} & \textbf{0.152} & \textbf{30.05} & \textbf{0.73} \\
\bottomrule
\end{tabular}
}
\caption{Quantitative results on MVHumanNet and DNA-Rendering. ↓ indicates lower is better while ↑ indicates higher is better. ReCamMaster* is the finetuned version using MVHumanNet.}
\label{tab:stacked_results}
\end{table}

\subsection{Quantitative and Qualitative Results}
\begin{figure*}
\centering
  \includegraphics[trim=0cm 0cm 0cm 0cm, clip=true,width=0.95\linewidth]{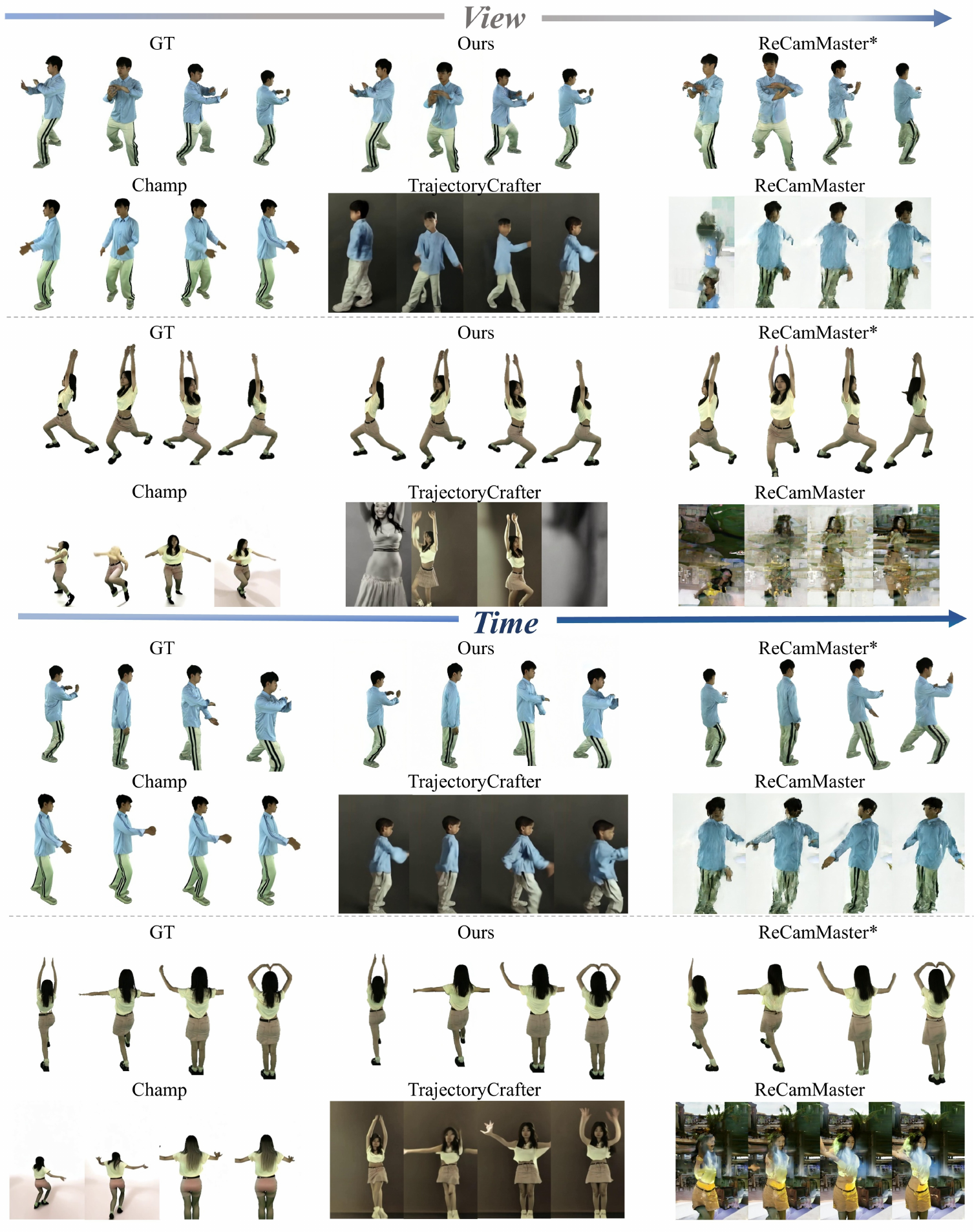}
  \caption{\textbf{Comparison with state-of-the-art methods tested on MVHumanNet dataset. ReCamMaster* is the finetuned version using MVHumanNet.} }
  \label{fig:mvhuman_result}
\end{figure*}

\begin{figure*}
\centering
  \includegraphics[trim=0cm 0cm 0cm 0cm, clip=true, width=0.95\linewidth]{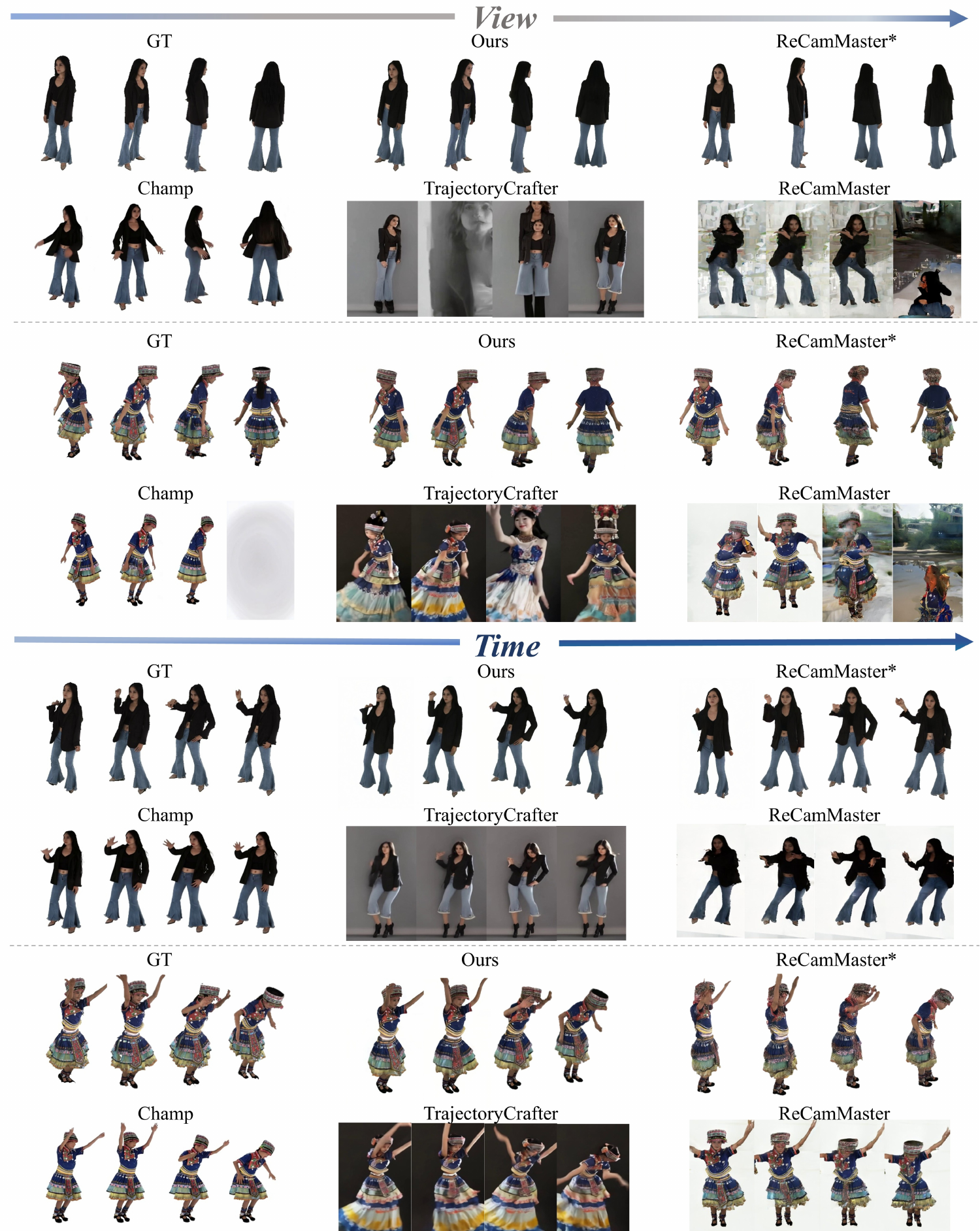}
  \caption{\textbf{Comparison with state-of-the-art methods tested on DNA-rendering dataset. ReCamMaster* is the finetuned version using MVHumanNet.}}

  \label{fig:dna_result}
\end{figure*}

\cref{tab:stacked_results} presents the quantitative results on two datasets, which show that existing models~\cite{zhu2024champ, yu2025trajectorycrafter, bai2025recammaster} are not good at this task. Our method is the first to achieve faithful and 360-degree synchronized multi-view synthesis from human-centric monocular video. We exhibit the qualitative comparisons using two datasets in \cref{fig:mvhuman_result} \cref{fig:dna_result}, respectively. It can be observed that \PIPENAME{} outperforms all baselines by an order of magnitude. Notably, our generated frontal videos are nearly pixel-aligned with the frontal ground truth, while \PIPENAME{} also produces consistent and reasonable back-view imagination. This is consistent with the reported FVD scores. Moreover, \PIPENAME{} accepts only frontal-view videos as input, while the backside clothing patterns are synthesized by the video diffusion model. Although discrepancies exist between the generated backside textures and the ground truth, the results remain reasonable and acceptable. Visually, both ReCamMaster and TrajectoryCrafter can only produce plausible frontal views while struggling to generate significant viewpoint changes in the video.
ReCamMaster*, the finetuned version model, shows improvements across all metrics. However, it remains deficient in fine-grained camera control and struggles with generalizing to out-of-distribution camera poses. This issue of leveraging implicit camera embedding is also highlighted in \citet{tang2025gaf}. Despite Champ \cite{zhu2024champ}, being adapted from an image-based model rather than a native video generation framework, struggles to preserve identity consistency during animation. Besides, these methods fail to maintain consistency across different viewpoints. In contrast, our method is capable of generating coherent and faithful 360-degree multi-view synthesis, even in challenging scenarios involving complex clothing. For additional visual results of in-the-wild performers, please refer to the supplementary video.

\subsection{Ablation Studies}
We ablate each component in \PIPENAME{} using MVHumanNet \cite{xiong2024mvhumannet}, DNA-Rendering \cite{2023dnarendering} and in-the-wild dataset.

\begin{table}[ht]
\vspace{-1mm}
\resizebox{\columnwidth}{!}{
\begin{tabular}{l|lllll}
\hline

Method              & \multicolumn{1}{c}{PSNR ↑} & \multicolumn{1}{c}{SSIM ↑} & \multicolumn{1}{c}{LPIPS ↓} & \multicolumn{1}{c}{FID ↓} & \multicolumn{1}{c}{FVD ↓} \\ \hline

w/o normal cond (A) &          15.61             &            0.858        &             0.165         &         36.60          &            0.837        \\
w/o sync module (B)        &       15.38              &         0.856             &          0.163          &       38.96              &          0.898          \\
w/o (A) \& w/o (B) &          15.21             &            0.850        &             0.169         &         39.13         &            1.06    \\ \hline
\textbf{Ours full} &
\textbf{15.63} & \textbf{0.861} & \textbf{0.152} & \textbf{30.05} & \textbf{0.73}                  \\ \hline
\end{tabular}
}
\caption{Ablation Studies on the whole framework.}
\label{tab:ablation}
\vspace{-4mm}
\end{table}

\begin{figure}[t]
    \centering
\includegraphics[width=\columnwidth]{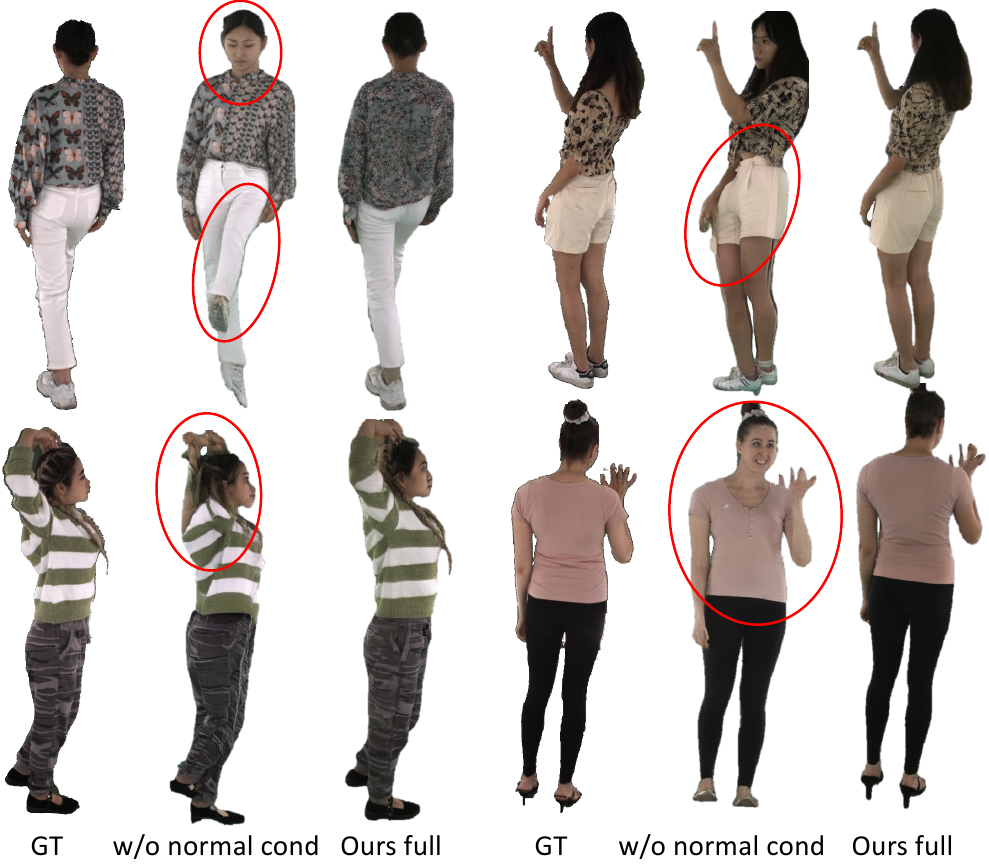}
    \caption{Our proposed camera-dependent normal condition assists the model in distinguishing between observed and unobserved condition information, resulting in a more accurate 360-degree synthesis.}
    \label{fig:ab-normalcond}

\end{figure}

\textbf{Camera-dependent normal condition.}
To demonstrate the effectiveness of our proposed conditioning signal, we conducted an experiment where this signal was omitted during the finetuning process. As shown in \cref{tab:ablation}, all metrics exhibit a noticeable degradation without camera-dependent normal condition. Furthermore, \cref{fig:ab-normalcond} illustrates these results more clearly. It can be observed that without the facilitation of our proposed condition signal, the model produces incorrect results due to the condition ambiguity, which indicates that the normal condition serves as a strong geometric cue, alleviating such errors. Notably, this can be regarded as a finetuned version of TrajectoryCrafter \cite{yu2025trajectorycrafter} on  MVHumanNet \cite{xiong2024mvhumannet} dataset. We emphasize that our customized design plays a crucial role in addressing this challenging problem.

\begin{figure}[h]
    \centering
\includegraphics[width=\columnwidth]{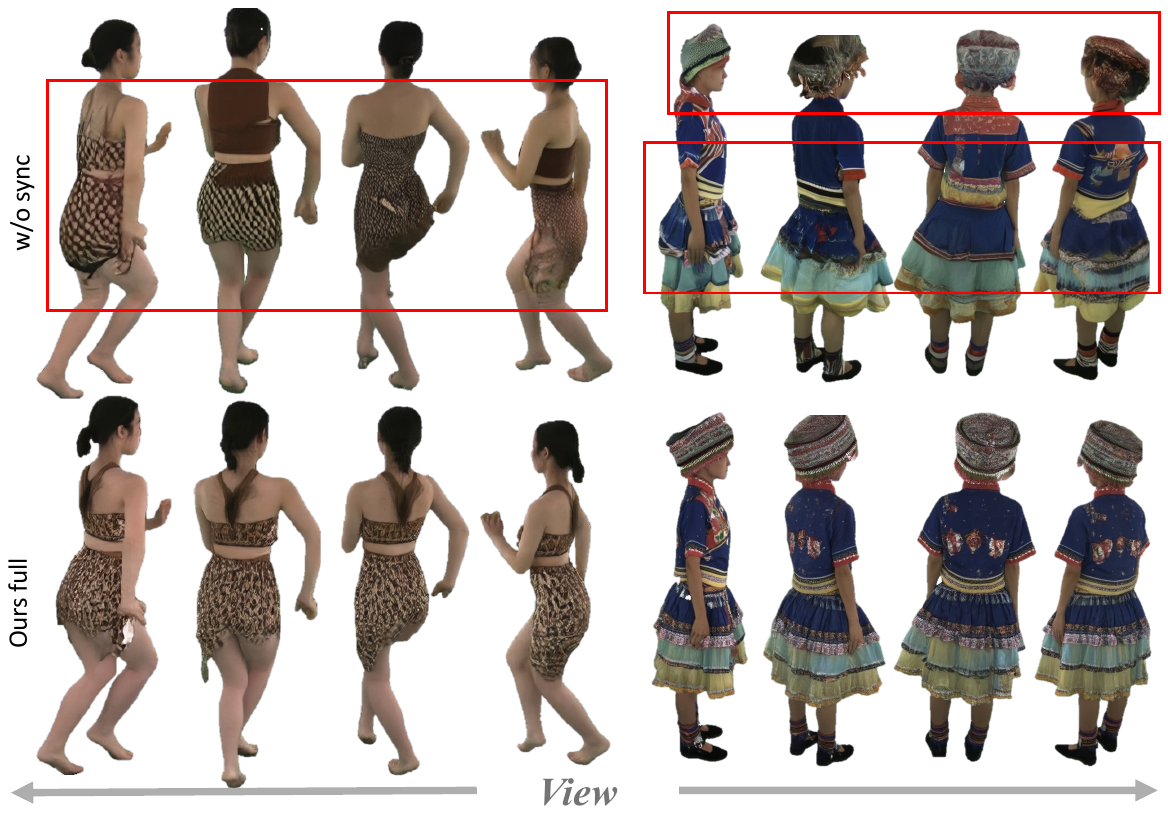}
    \caption{The syncronization attention largely enhance the generation consistency across views.}
    \label{fig:ab-sync}
\vspace{-2mm}

\end{figure}

\textbf{Sync module.} As discussed in \cref{sec:mv-vdm}, most existing camera-controllable video diffusion models face challenges maintaining consistency across different views. To address this issue, we implement synchronization attention to improve 4D view consistency. As illustrated in \cref{fig:ab-sync}, the incorporation of the view-sync module results in a more consistent and visually enhanced appearance. Furthermore, the synchronization operation can also enhance the quantitative performance.

\begin{figure}[h]
    \centering
\includegraphics[width=\columnwidth]{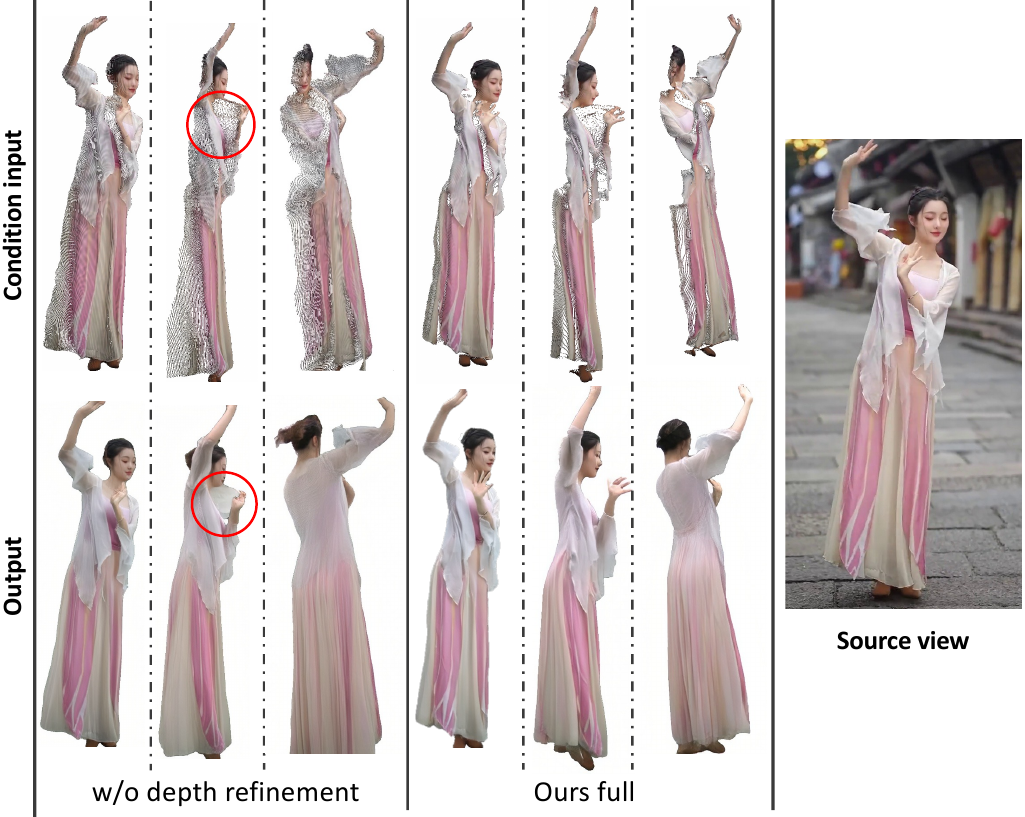}
    \caption{The initial estimated point clouds contain floaters near the edges of the character, leading to bad guidance to the video diffusion model. In contrast, our method achieves clean estimations and yields pleasing results.}
    \label{fig:ab-depthrefine}
\end{figure}

\textbf{Depth refinement.} We evaluate the effectiveness of the depth refinement process on the in-the-wild data by replacing depth with the initial estimation from MegaSaM \cite{li2024_megasam}. As exhibited in \cref{fig:ab-depthrefine}, it is evident that depth fidelity significantly influences the final results. Inaccurate depth maps result in noisy warping, and this issue intensifies with increasing viewpoint changes (from left to right). The model generates unnatural body appearances due to floaters in the condition signals near the human body. In contrast, our integrated depth refinement process mitigates these floaters caused by inaccurate monocular depth estimations, generating clean point clouds. We achieve high-quality generation outcomes with clean geometric cue conditions.

\textbf{Sampling steps.} We also show the influence of sampling steps in \cref{tab:samplesteps}. Reducing the sampling steps leads to poorer performance, particularly in FID. 25-50 denoising steps strike a balance between quality and cost.

\begin{table}[htbp]
\centering
\begin{tabular}{l | c c c c c}
\toprule
Steps & PSNR ↑ & SSIM ↑ & LPIPS ↑ & FID ↓ & FVD ↓ \\
\midrule
\multicolumn{6}{c}{\textbf{MVHumanNet~\cite{xiong2024mvhumannet}}} \\
\midrule
5  & 24.90 & 0.931 & 0.078 & 55.54 & 0.14 \\
10 & 24.65 & 0.929 & 0.074 & 43.43 & 0.13 \\
25 & 24.40 & 0.927 & 0.069 & 30.26 & 0.12 \\
50 & 24.35 & 0.926 & 0.066 & 24.47 & 0.12 \\
\midrule
\multicolumn{6}{c}{\textbf{DNA-Rendering~\cite{2023dnarendering}}} \\
\midrule
5  & 15.72 & 0.864 & 0.166 & 54.85 & 0.74 \\
10 & 15.65 & 0.862 & 0.161 & 45.00 & 0.74 \\
25 & 15.63 & 0.861 & 0.155 & 34.97 & 0.74 \\
50 & 15.63 & 0.861 & 0.152 & 30.05 & 0.73 \\
\bottomrule
\end{tabular}
\caption{Performance under different sampling steps.}
\label{tab:samplesteps}
\end{table}

\subsection{Application}
\begin{figure}[h]
    \centering
\includegraphics[width=\columnwidth]{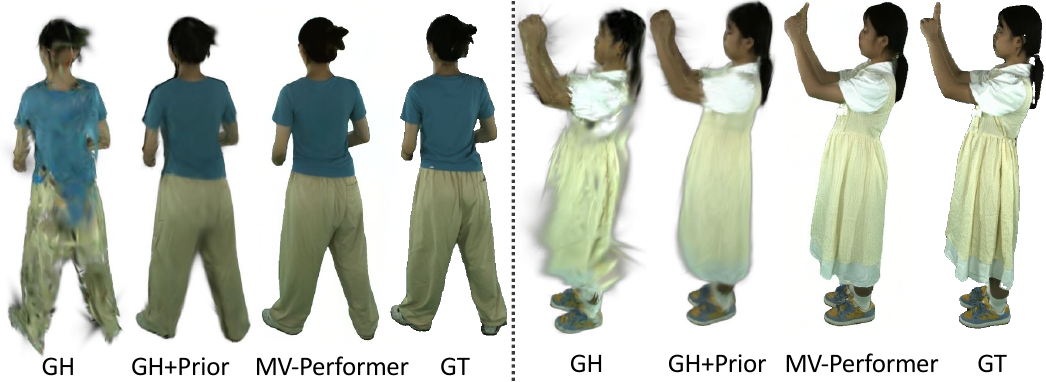}
    \caption{Using \PIPENAME{} as a generative prior. ``GH'' means GauHuman \cite{hu2024gauhuman}}
    \label{fig:app-prior}
\end{figure}

An application of generative novel view synthesizers is to serve as generative priors \cite{tang2025gaf, yu2024viewcrafter, liu2023zero, shi2023zero123++, jiang2024animate3d}. We show that \PIPENAME{} could potentially act as a prior for monocular avatar reconstruction. Without loss of generality, we add the comparison with GauHuman \cite{hu2024gauhuman} on MVHumanNet \cite{xiong2024mvhumannet}. Specifically, we use \PIPENAME{} to generate two side-view and one back-view videos from frontal view videos as priors. We combine them with original frontal view videos to train GauHuman \cite{hu2024gauhuman}. As shown in \cref{fig:app-prior} and \cref{tab:mvp_prior}, due to limited observations, GauHuman \cite{hu2024gauhuman} produces strong artifacts when viewed from the rear, resulting in poorer results. After incorporating the prior, we observe performance improvements across all metrics, reducing the artifacts behind the performers. \cref{fig:app-prior} and \cref{tab:mvp_prior} also reveal the potential of directly using the video diffusion model to perform 4D novel view synthesis.

\begin{table}[htbp]
\centering
\begin{tabular}{l | c c c c c}
\hline
Methods & PSNR↑ & SSIM↑ & LPIPS↑ & FID↓ & FVD↓ \\ \hline
GauHuman & 18.63 & 0.866 & 0.179 & 129.35 & 5.96 \\
GauHuman+Prior & 20.97 & 0.901 & 0.146 & 60.02 & 1.81 \\
MV-Performer & 24.35 & 0.926 & 0.066 & 24.47 & 0.12 \\
\hline
\end{tabular}
\label{tab:mvp_prior}
\caption{We validate the effectiveness of prior on MVHumanNet.}
\vspace{-2mm}
\end{table}

\section{Limitations}
For the training process, despite the robust VAE offered by WAN2.1 \cite{wan2025}, preserving face region details remains challenging due to reconstruction errors, which limit the upper bounds of the human generation quality. For inference, \PIPENAME{} essentially counts on the stability of the depth estimation methods \cite{piccinelli2025unidepthv2,li2024_megasam}. Our generated results would fail when faced with poor depth estimation. However, this problem could be solved by finetuning the depth estimation model with the metric human depth in MVHumanNet++~\cite{li2025mvhumannet++}. Moreover, the video diffusion model generally requires multi-step denoising during inference, resulting in relatively high computational overhead and slow inference speed. Distilling \PIPENAME{} into a smaller and one-step denoising version \cite{wang2025videoscene} is a promising direction toward practical application. \PIPENAME{} may degrade in quality for untrained origin and certain skin tones, which is limited by the potential bias in WAN2.1 and the existing dataset. Finally, limited by the computational resource, we can only conduct experiments on the 1.3B version of WAN2.1 \cite{wan2025}.

\section{Conclusion}

In this paper, we present \PIPENAME, a novel framework for 360-degree human-centric novel view synthesis from monocular full-body videos. To address the limitations of existing warping-based methods, which often struggle with significant viewpoint changes, we introduce a camera-dependent normal map geometric condition signal. This approach effectively resolves the ambiguity between seen and unseen regions of the input human performer.  Furthermore, we proposed a robust inference procedure to handle in-the-wild videos, significantly reducing artifacts caused by imperfect monocular depth estimation.
Benefiting from the aforementioned design, our multi-view human-centric video diffusion model ensures temporal and geometric consistency across synthesized viewpoints.
Extensive experiments on three datasets validate that \PIPENAME{} outperforms the existing camera-controllable video diffusion model, establishing a strong model for 4D human-centric novel view synthesis. Our framework opens new possibilities for immersive VR/AR, free-viewpoint video, and synthetic data generation, which will benefit numerous downstream tasks.

\begin{acks}
The work was supported in part by NSFC with Grant No. 62293482, the Basic Research Project No. HZQB-KCZYZ-2021067 of Hetao Shenzhen-HK S\&T Cooperation Zone, Guangdong Provincial Outstanding Youth Fund (No. 2023B1515020055), by Shenzhen Science and Technology Program No. JCYJ20220530143604010, NSFC No. 62172348, the Shenzhen Outstanding Talents Training Fund 202002, the Guangdong Research Projects No. 2017ZT07X152 and No. 2019CX01X104, the Guangdong Provincial Key Laboratory of Future Networks of Intelligence (Grant No. 2022B1212010001), and the Shenzhen Key Laboratory of Big Data and Artificial Intelligence (Grant No. SYSPG20241211173853027). 

\end{acks}


\clearpage
\bibliographystyle{ACM-Reference-Format}
\bibliography{bibliography}


\begin{thebibliography}{154}


\ifx \showCODEN    \undefined \def \showCODEN     #1{\unskip}     \fi
\ifx \showISBNx    \undefined \def \showISBNx     #1{\unskip}     \fi
\ifx \showISBNxiii \undefined \def \showISBNxiii  #1{\unskip}     \fi
\ifx \showISSN     \undefined \def \showISSN      #1{\unskip}     \fi
\ifx \showLCCN     \undefined \def \showLCCN      #1{\unskip}     \fi
\ifx \shownote     \undefined \def \shownote      #1{#1}          \fi
\ifx \showarticletitle \undefined \def \showarticletitle #1{#1}   \fi
\ifx \showURL      \undefined \def \showURL       {\relax}        \fi
\providecommand\bibfield[2]{#2}
\providecommand\bibinfo[2]{#2}
\providecommand\natexlab[1]{#1}
\providecommand\showeprint[2][]{arXiv:#2}

\bibitem[Avidan and Shashua(1997)]%
        {avidan1997novel}
\bibfield{author}{\bibinfo{person}{Shai Avidan} {and} \bibinfo{person}{Amnon Shashua}.} \bibinfo{year}{1997}\natexlab{}.
\newblock \showarticletitle{Novel view synthesis in tensor space}. In \bibinfo{booktitle}{\emph{Proceedings of IEEE Computer Society Conference on Computer Vision and Pattern Recognition}}. IEEE, \bibinfo{pages}{1034--1040}.
\newblock


\bibitem[Bahmani et~al\mbox{.}(2024)]%
        {bahmani20244d}
\bibfield{author}{\bibinfo{person}{Sherwin Bahmani}, \bibinfo{person}{Ivan Skorokhodov}, \bibinfo{person}{Victor Rong}, \bibinfo{person}{Gordon Wetzstein}, \bibinfo{person}{Leonidas Guibas}, \bibinfo{person}{Peter Wonka}, \bibinfo{person}{Sergey Tulyakov}, \bibinfo{person}{Jeong~Joon Park}, \bibinfo{person}{Andrea Tagliasacchi}, {and} \bibinfo{person}{David~B Lindell}.} \bibinfo{year}{2024}\natexlab{}.
\newblock \showarticletitle{4d-fy: Text-to-4d generation using hybrid score distillation sampling}. In \bibinfo{booktitle}{\emph{Proceedings of the IEEE/CVF Conference on Computer Vision and Pattern Recognition}}. \bibinfo{pages}{7996--8006}.
\newblock


\bibitem[Bai et~al\mbox{.}(2025)]%
        {bai2025recammaster}
\bibfield{author}{\bibinfo{person}{Jianhong Bai}, \bibinfo{person}{Menghan Xia}, \bibinfo{person}{Xiao Fu}, \bibinfo{person}{Xintao Wang}, \bibinfo{person}{Lianrui Mu}, \bibinfo{person}{Jinwen Cao}, \bibinfo{person}{Zuozhu Liu}, \bibinfo{person}{Haoji Hu}, \bibinfo{person}{Xiang Bai}, \bibinfo{person}{Pengfei Wan}, {and} \bibinfo{person}{Di Zhang}.} \bibinfo{year}{2025}\natexlab{}.
\newblock \bibinfo{title}{ReCamMaster: Camera-Controlled Generative Rendering from A Single Video}.
\newblock
\showeprint[arxiv]{2503.11647}~[cs.CV]
\urldef\tempurl%
\url{https://arxiv.org/abs/2503.11647}
\showURL{%
\tempurl}


\bibitem[Bai et~al\mbox{.}(2024)]%
        {bai2024syncammaster}
\bibfield{author}{\bibinfo{person}{Jianhong Bai}, \bibinfo{person}{Menghan Xia}, \bibinfo{person}{Xintao Wang}, \bibinfo{person}{Ziyang Yuan}, \bibinfo{person}{Xiao Fu}, \bibinfo{person}{Zuozhu Liu}, \bibinfo{person}{Haoji Hu}, \bibinfo{person}{Pengfei Wan}, {and} \bibinfo{person}{Di Zhang}.} \bibinfo{year}{2024}\natexlab{}.
\newblock \bibinfo{title}{SynCamMaster: Synchronizing Multi-Camera Video Generation from Diverse Viewpoints}.
\newblock
\showeprint[arxiv]{2412.07760}~[cs.CV]
\urldef\tempurl%
\url{https://arxiv.org/abs/2412.07760}
\showURL{%
\tempurl}


\bibitem[Bian et~al\mbox{.}(2025)]%
        {bian2025gs}
\bibfield{author}{\bibinfo{person}{Weikang Bian}, \bibinfo{person}{Zhaoyang Huang}, \bibinfo{person}{Xiaoyu Shi}, \bibinfo{person}{Yijin Li}, \bibinfo{person}{Fu-Yun Wang}, {and} \bibinfo{person}{Hongsheng Li}.} \bibinfo{year}{2025}\natexlab{}.
\newblock \showarticletitle{GS-DiT: Advancing Video Generation with Pseudo 4D Gaussian Fields through Efficient Dense 3D Point Tracking}.
\newblock \bibinfo{journal}{\emph{arXiv preprint arXiv:2501.02690}} (\bibinfo{year}{2025}).
\newblock


\bibitem[Blattmann et~al\mbox{.}(2023a)]%
        {blattmann2023stable}
\bibfield{author}{\bibinfo{person}{Andreas Blattmann}, \bibinfo{person}{Tim Dockhorn}, \bibinfo{person}{Sumith Kulal}, \bibinfo{person}{Daniel Mendelevitch}, \bibinfo{person}{Maciej Kilian}, \bibinfo{person}{Dominik Lorenz}, \bibinfo{person}{Yam Levi}, \bibinfo{person}{Zion English}, \bibinfo{person}{Vikram Voleti}, \bibinfo{person}{Adam Letts}, {et~al\mbox{.}}} \bibinfo{year}{2023}\natexlab{a}.
\newblock \showarticletitle{Stable video diffusion: Scaling latent video diffusion models to large datasets}.
\newblock \bibinfo{journal}{\emph{arXiv preprint arXiv:2311.15127}} (\bibinfo{year}{2023}).
\newblock


\bibitem[Blattmann et~al\mbox{.}(2023b)]%
        {blattmann2023align}
\bibfield{author}{\bibinfo{person}{Andreas Blattmann}, \bibinfo{person}{Robin Rombach}, \bibinfo{person}{Huan Ling}, \bibinfo{person}{Tim Dockhorn}, \bibinfo{person}{Seung~Wook Kim}, \bibinfo{person}{Sanja Fidler}, {and} \bibinfo{person}{Karsten Kreis}.} \bibinfo{year}{2023}\natexlab{b}.
\newblock \showarticletitle{Align your latents: High-resolution video synthesis with latent diffusion models}. In \bibinfo{booktitle}{\emph{Proceedings of the IEEE/CVF conference on computer vision and pattern recognition}}. \bibinfo{pages}{22563--22575}.
\newblock


\bibitem[Cao and Johnson(2023)]%
        {cao2023hexplane}
\bibfield{author}{\bibinfo{person}{Ang Cao} {and} \bibinfo{person}{Justin Johnson}.} \bibinfo{year}{2023}\natexlab{}.
\newblock \showarticletitle{Hexplane: A fast representation for dynamic scenes}. In \bibinfo{booktitle}{\emph{Proceedings of the IEEE/CVF Conference on Computer Vision and Pattern Recognition}}. \bibinfo{pages}{130--141}.
\newblock


\bibitem[Cao et~al\mbox{.}(2022)]%
        {bini2022cao}
\bibfield{author}{\bibinfo{person}{Xu Cao}, \bibinfo{person}{Hiroaki Santo}, \bibinfo{person}{Boxin Shi}, \bibinfo{person}{Fumio Okura}, {and} \bibinfo{person}{Yasuyuki Matsushita}.} \bibinfo{year}{2022}\natexlab{}.
\newblock \showarticletitle{Bilateral Normal Integration}. In \bibinfo{booktitle}{\emph{ECCV}}.
\newblock


\bibitem[Chaurasia et~al\mbox{.}(2011)]%
        {chaurasia2011silhouette}
\bibfield{author}{\bibinfo{person}{Gaurav Chaurasia}, \bibinfo{person}{Olga Sorkine}, {and} \bibinfo{person}{George Drettakis}.} \bibinfo{year}{2011}\natexlab{}.
\newblock \showarticletitle{Silhouette-Aware Warping for Image-Based Rendering}. In \bibinfo{booktitle}{\emph{Computer Graphics Forum}}, Vol.~\bibinfo{volume}{30}. Wiley Online Library, \bibinfo{pages}{1223--1232}.
\newblock


\bibitem[Chen et~al\mbox{.}(2022a)]%
        {Chen2022ECCV}
\bibfield{author}{\bibinfo{person}{Anpei Chen}, \bibinfo{person}{Zexiang Xu}, \bibinfo{person}{Andreas Geiger}, \bibinfo{person}{Jingyi Yu}, {and} \bibinfo{person}{Hao Su}.} \bibinfo{year}{2022}\natexlab{a}.
\newblock \showarticletitle{TensoRF: Tensorial Radiance Fields}. In \bibinfo{booktitle}{\emph{European Conference on Computer Vision (ECCV)}}.
\newblock


\bibitem[Chen et~al\mbox{.}(2021)]%
        {mvsnerf}
\bibfield{author}{\bibinfo{person}{Anpei Chen}, \bibinfo{person}{Zexiang Xu}, \bibinfo{person}{Fuqiang Zhao}, \bibinfo{person}{Xiaoshuai Zhang}, \bibinfo{person}{Fanbo Xiang}, \bibinfo{person}{Jingyi Yu}, {and} \bibinfo{person}{Hao Su}.} \bibinfo{year}{2021}\natexlab{}.
\newblock \showarticletitle{Mvsnerf: Fast generalizable radiance field reconstruction from multi-view stereo}. In \bibinfo{booktitle}{\emph{Proceedings of the IEEE/CVF International Conference on Computer Vision}}. \bibinfo{pages}{14124--14133}.
\newblock


\bibitem[Chen et~al\mbox{.}(2023)]%
        {chen2023videocrafter1}
\bibfield{author}{\bibinfo{person}{Haoxin Chen}, \bibinfo{person}{Menghan Xia}, \bibinfo{person}{Yingqing He}, \bibinfo{person}{Yong Zhang}, \bibinfo{person}{Xiaodong Cun}, \bibinfo{person}{Shaoshu Yang}, \bibinfo{person}{Jinbo Xing}, \bibinfo{person}{Yaofang Liu}, \bibinfo{person}{Qifeng Chen}, \bibinfo{person}{Xintao Wang}, \bibinfo{person}{Chao Weng}, {and} \bibinfo{person}{Ying Shan}.} \bibinfo{year}{2023}\natexlab{}.
\newblock \bibinfo{title}{VideoCrafter1: Open Diffusion Models for High-Quality Video Generation}.
\newblock
\showeprint[arxiv]{2310.19512}~[cs.CV]


\bibitem[Chen et~al\mbox{.}(2024b)]%
        {chen2024videocrafter2}
\bibfield{author}{\bibinfo{person}{Haoxin Chen}, \bibinfo{person}{Yong Zhang}, \bibinfo{person}{Xiaodong Cun}, \bibinfo{person}{Menghan Xia}, \bibinfo{person}{Xintao Wang}, \bibinfo{person}{Chao Weng}, {and} \bibinfo{person}{Ying Shan}.} \bibinfo{year}{2024}\natexlab{b}.
\newblock \bibinfo{title}{VideoCrafter2: Overcoming Data Limitations for High-Quality Video Diffusion Models}.
\newblock
\showeprint[arxiv]{2401.09047}~[cs.CV]


\bibitem[Chen et~al\mbox{.}(2025)]%
        {chen2025taoavatar}
\bibfield{author}{\bibinfo{person}{Jianchuan Chen}, \bibinfo{person}{Jingchuan Hu}, \bibinfo{person}{Gaige Wang}, \bibinfo{person}{Zhonghua Jiang}, \bibinfo{person}{Tiansong Zhou}, \bibinfo{person}{Zhiwen Chen}, {and} \bibinfo{person}{Chengfei Lv}.} \bibinfo{year}{2025}\natexlab{}.
\newblock \showarticletitle{TaoAvatar: Real-Time Lifelike Full-Body Talking Avatars for Augmented Reality via 3D Gaussian Splatting}. In \bibinfo{booktitle}{\emph{Proceedings of the Computer Vision and Pattern Recognition Conference}}. \bibinfo{pages}{10723--10734}.
\newblock


\bibitem[Chen et~al\mbox{.}(2022b)]%
        {chen2022geometry}
\bibfield{author}{\bibinfo{person}{Mingfei Chen}, \bibinfo{person}{Jianfeng Zhang}, \bibinfo{person}{Xiangyu Xu}, \bibinfo{person}{Lijuan Liu}, \bibinfo{person}{Yujun Cai}, \bibinfo{person}{Jiashi Feng}, {and} \bibinfo{person}{Shuicheng Yan}.} \bibinfo{year}{2022}\natexlab{b}.
\newblock \showarticletitle{Geometry-guided progressive nerf for generalizable and efficient neural human rendering}. In \bibinfo{booktitle}{\emph{European Conference on Computer Vision}}. Springer, \bibinfo{pages}{222--239}.
\newblock


\bibitem[Chen and Williams(2023)]%
        {chen2023view}
\bibfield{author}{\bibinfo{person}{Shenchang~Eric Chen} {and} \bibinfo{person}{Lance Williams}.} \bibinfo{year}{2023}\natexlab{}.
\newblock \showarticletitle{View interpolation for image synthesis}.
\newblock In \bibinfo{booktitle}{\emph{Seminal Graphics Papers: Pushing the Boundaries, Volume 2}}. \bibinfo{pages}{423--432}.
\newblock


\bibitem[Chen et~al\mbox{.}(2024a)]%
        {chen2024mvsplat}
\bibfield{author}{\bibinfo{person}{Yuedong Chen}, \bibinfo{person}{Haofei Xu}, \bibinfo{person}{Chuanxia Zheng}, \bibinfo{person}{Bohan Zhuang}, \bibinfo{person}{Marc Pollefeys}, \bibinfo{person}{Andreas Geiger}, \bibinfo{person}{Tat-Jen Cham}, {and} \bibinfo{person}{Jianfei Cai}.} \bibinfo{year}{2024}\natexlab{a}.
\newblock \showarticletitle{Mvsplat: Efficient 3d gaussian splatting from sparse multi-view images}. In \bibinfo{booktitle}{\emph{European Conference on Computer Vision}}. Springer, \bibinfo{pages}{370--386}.
\newblock


\bibitem[Chen et~al\mbox{.}(2024c)]%
        {chen2024meshavatar}
\bibfield{author}{\bibinfo{person}{Yushuo Chen}, \bibinfo{person}{Zerong Zheng}, \bibinfo{person}{Zhe Li}, \bibinfo{person}{Chao Xu}, {and} \bibinfo{person}{Yebin Liu}.} \bibinfo{year}{2024}\natexlab{c}.
\newblock \showarticletitle{Meshavatar: Learning high-quality triangular human avatars from multi-view videos}. In \bibinfo{booktitle}{\emph{European Conference on Computer Vision}}. Springer, \bibinfo{pages}{250--269}.
\newblock


\bibitem[Cheng et~al\mbox{.}(2023)]%
        {2023dnarendering}
\bibfield{author}{\bibinfo{person}{Wei Cheng}, \bibinfo{person}{Ruixiang Chen}, \bibinfo{person}{Wanqi Yin}, \bibinfo{person}{Siming Fan}, \bibinfo{person}{Keyu Chen}, \bibinfo{person}{Honglin He}, \bibinfo{person}{Huiwen Luo}, \bibinfo{person}{Zhongang Cai}, \bibinfo{person}{Jingbo Wang}, \bibinfo{person}{Yang Gao}, \bibinfo{person}{Zhengming Yu}, \bibinfo{person}{Zhengyu Lin}, \bibinfo{person}{Daxuan Ren}, \bibinfo{person}{Lei Yang}, \bibinfo{person}{Ziwei Liu}, \bibinfo{person}{Chen~Change Loy}, \bibinfo{person}{Chen Qian}, \bibinfo{person}{Wayne Wu}, \bibinfo{person}{Dahua Lin}, \bibinfo{person}{Bo Dai}, {and} \bibinfo{person}{Kwan-Yee Lin}.} \bibinfo{year}{2023}\natexlab{}.
\newblock \showarticletitle{DNA-Rendering: A Diverse Neural Actor Repository for High-Fidelity Human-centric Rendering}.
\newblock \bibinfo{journal}{\emph{arXiv preprint}}  \bibinfo{volume}{arXiv:2307.10173} (\bibinfo{year}{2023}).
\newblock


\bibitem[Duan et~al\mbox{.}(2024)]%
        {duan20244d}
\bibfield{author}{\bibinfo{person}{Yuanxing Duan}, \bibinfo{person}{Fangyin Wei}, \bibinfo{person}{Qiyu Dai}, \bibinfo{person}{Yuhang He}, \bibinfo{person}{Wenzheng Chen}, {and} \bibinfo{person}{Baoquan Chen}.} \bibinfo{year}{2024}\natexlab{}.
\newblock \showarticletitle{4d-rotor gaussian splatting: towards efficient novel view synthesis for dynamic scenes}. In \bibinfo{booktitle}{\emph{ACM SIGGRAPH 2024 Conference Papers}}. \bibinfo{pages}{1--11}.
\newblock


\bibitem[Esser et~al\mbox{.}(2024)]%
        {esser2024scaling}
\bibfield{author}{\bibinfo{person}{Patrick Esser}, \bibinfo{person}{Sumith Kulal}, \bibinfo{person}{Andreas Blattmann}, \bibinfo{person}{Rahim Entezari}, \bibinfo{person}{Jonas M{\"u}ller}, \bibinfo{person}{Harry Saini}, \bibinfo{person}{Yam Levi}, \bibinfo{person}{Dominik Lorenz}, \bibinfo{person}{Axel Sauer}, \bibinfo{person}{Frederic Boesel}, {et~al\mbox{.}}} \bibinfo{year}{2024}\natexlab{}.
\newblock \showarticletitle{Scaling rectified flow transformers for high-resolution image synthesis}. In \bibinfo{booktitle}{\emph{Forty-first international conference on machine learning}}.
\newblock


\bibitem[Fridovich-Keil et~al\mbox{.}(2023)]%
        {fridovich2023k}
\bibfield{author}{\bibinfo{person}{Sara Fridovich-Keil}, \bibinfo{person}{Giacomo Meanti}, \bibinfo{person}{Frederik~Rahb{\ae}k Warburg}, \bibinfo{person}{Benjamin Recht}, {and} \bibinfo{person}{Angjoo Kanazawa}.} \bibinfo{year}{2023}\natexlab{}.
\newblock \showarticletitle{K-planes: Explicit radiance fields in space, time, and appearance}. In \bibinfo{booktitle}{\emph{Proceedings of the IEEE/CVF Conference on Computer Vision and Pattern Recognition}}. \bibinfo{pages}{12479--12488}.
\newblock


\bibitem[Furukawa et~al\mbox{.}(2015)]%
        {furukawa2015multi}
\bibfield{author}{\bibinfo{person}{Yasutaka Furukawa}, \bibinfo{person}{Carlos Hern{\'a}ndez}, {et~al\mbox{.}}} \bibinfo{year}{2015}\natexlab{}.
\newblock \showarticletitle{Multi-view stereo: A tutorial}.
\newblock \bibinfo{journal}{\emph{Foundations and trends{\textregistered} in Computer Graphics and Vision}} \bibinfo{volume}{9}, \bibinfo{number}{1-2} (\bibinfo{year}{2015}), \bibinfo{pages}{1--148}.
\newblock


\bibitem[Glasbey and Mardia(1998)]%
        {glasbey1998review}
\bibfield{author}{\bibinfo{person}{Chris~A Glasbey} {and} \bibinfo{person}{Kantilal~Vardichand Mardia}.} \bibinfo{year}{1998}\natexlab{}.
\newblock \showarticletitle{A review of image-warping methods}.
\newblock \bibinfo{journal}{\emph{Journal of applied statistics}} \bibinfo{volume}{25}, \bibinfo{number}{2} (\bibinfo{year}{1998}), \bibinfo{pages}{155--171}.
\newblock


\bibitem[Gu et~al\mbox{.}(2025)]%
        {gu2025das}
\bibfield{author}{\bibinfo{person}{Zekai Gu}, \bibinfo{person}{Rui Yan}, \bibinfo{person}{Jiahao Lu}, \bibinfo{person}{Peng Li}, \bibinfo{person}{Zhiyang Dou}, \bibinfo{person}{Chenyang Si}, \bibinfo{person}{Zhen Dong}, \bibinfo{person}{Qifeng Liu}, \bibinfo{person}{Cheng Lin}, \bibinfo{person}{Ziwei Liu}, \bibinfo{person}{Wenping Wang}, {and} \bibinfo{person}{Yuan Liu}.} \bibinfo{year}{2025}\natexlab{}.
\newblock \showarticletitle{Diffusion as Shader: 3D-aware Video Diffusion for Versatile Video Generation Control}.
\newblock \bibinfo{journal}{\emph{arXiv preprint arXiv:2501.03847}} (\bibinfo{year}{2025}).
\newblock


\bibitem[Guo et~al\mbox{.}(2023)]%
        {guo2023vid2avatar}
\bibfield{author}{\bibinfo{person}{Chen Guo}, \bibinfo{person}{Tianjian Jiang}, \bibinfo{person}{Xu Chen}, \bibinfo{person}{Jie Song}, {and} \bibinfo{person}{Otmar Hilliges}.} \bibinfo{year}{2023}\natexlab{}.
\newblock \showarticletitle{Vid2avatar: 3d avatar reconstruction from videos in the wild via self-supervised scene decomposition}. In \bibinfo{booktitle}{\emph{Proceedings of the IEEE/CVF Conference on Computer Vision and Pattern Recognition}}. \bibinfo{pages}{12858--12868}.
\newblock


\bibitem[He et~al\mbox{.}({[n.\,d.]})]%
        {hecameractrl}
\bibfield{author}{\bibinfo{person}{Hao He}, \bibinfo{person}{Yinghao Xu}, \bibinfo{person}{Yuwei Guo}, \bibinfo{person}{Gordon Wetzstein}, \bibinfo{person}{Bo Dai}, \bibinfo{person}{Hongsheng Li}, {and} \bibinfo{person}{Ceyuan Yang}.} \bibinfo{year}{[n.\,d.]}\natexlab{}.
\newblock \showarticletitle{CameraCtrl: Enabling Camera Control for Video Diffusion Models}. In \bibinfo{booktitle}{\emph{The Thirteenth International Conference on Learning Representations}}.
\newblock


\bibitem[He et~al\mbox{.}(2024)]%
        {he2024cameractrl}
\bibfield{author}{\bibinfo{person}{Hao He}, \bibinfo{person}{Yinghao Xu}, \bibinfo{person}{Yuwei Guo}, \bibinfo{person}{Gordon Wetzstein}, \bibinfo{person}{Bo Dai}, \bibinfo{person}{Hongsheng Li}, {and} \bibinfo{person}{Ceyuan Yang}.} \bibinfo{year}{2024}\natexlab{}.
\newblock \showarticletitle{Cameractrl: Enabling camera control for text-to-video generation}.
\newblock \bibinfo{journal}{\emph{arXiv preprint arXiv:2404.02101}} (\bibinfo{year}{2024}).
\newblock


\bibitem[He et~al\mbox{.}(2022)]%
        {he2022lvdm}
\bibfield{author}{\bibinfo{person}{Yingqing He}, \bibinfo{person}{Tianyu Yang}, \bibinfo{person}{Yong Zhang}, \bibinfo{person}{Ying Shan}, {and} \bibinfo{person}{Qifeng Chen}.} \bibinfo{year}{2022}\natexlab{}.
\newblock \showarticletitle{Latent Video Diffusion Models for High-Fidelity Long Video Generation}.
\newblock  (\bibinfo{year}{2022}).
\newblock
\showeprint[arxiv]{2211.13221}~[cs.CV]


\bibitem[Heusel et~al\mbox{.}(2017)]%
        {heusel2017gans}
\bibfield{author}{\bibinfo{person}{Martin Heusel}, \bibinfo{person}{Hubert Ramsauer}, \bibinfo{person}{Thomas Unterthiner}, \bibinfo{person}{Bernhard Nessler}, {and} \bibinfo{person}{Sepp Hochreiter}.} \bibinfo{year}{2017}\natexlab{}.
\newblock \showarticletitle{Gans trained by a two time-scale update rule converge to a local nash equilibrium}.
\newblock \bibinfo{journal}{\emph{Advances in neural information processing systems}}  \bibinfo{volume}{30} (\bibinfo{year}{2017}).
\newblock


\bibitem[Hilsmann et~al\mbox{.}(2020)]%
        {hilsmann2020going}
\bibfield{author}{\bibinfo{person}{Anna Hilsmann}, \bibinfo{person}{Philipp Fechteler}, \bibinfo{person}{Wieland Morgenstern}, \bibinfo{person}{Wolfgang Paier}, \bibinfo{person}{Ingo Feldmann}, \bibinfo{person}{Oliver Schreer}, {and} \bibinfo{person}{Peter Eisert}.} \bibinfo{year}{2020}\natexlab{}.
\newblock \showarticletitle{Going beyond free viewpoint: creating animatable volumetric video of human performances}.
\newblock \bibinfo{journal}{\emph{IET Computer Vision}} \bibinfo{volume}{14}, \bibinfo{number}{6} (\bibinfo{year}{2020}), \bibinfo{pages}{350--358}.
\newblock


\bibitem[Ho et~al\mbox{.}(2020)]%
        {ho2020denoising}
\bibfield{author}{\bibinfo{person}{Jonathan Ho}, \bibinfo{person}{Ajay Jain}, {and} \bibinfo{person}{Pieter Abbeel}.} \bibinfo{year}{2020}\natexlab{}.
\newblock \showarticletitle{Denoising diffusion probabilistic models}.
\newblock \bibinfo{journal}{\emph{Advances in neural information processing systems}}  \bibinfo{volume}{33} (\bibinfo{year}{2020}), \bibinfo{pages}{6840--6851}.
\newblock


\bibitem[Hong et~al\mbox{.}(2022)]%
        {hong2022cogvideo}
\bibfield{author}{\bibinfo{person}{Wenyi Hong}, \bibinfo{person}{Ming Ding}, \bibinfo{person}{Wendi Zheng}, \bibinfo{person}{Xinghan Liu}, {and} \bibinfo{person}{Jie Tang}.} \bibinfo{year}{2022}\natexlab{}.
\newblock \showarticletitle{Cogvideo: Large-scale pretraining for text-to-video generation via transformers}.
\newblock \bibinfo{journal}{\emph{arXiv preprint arXiv:2205.15868}} (\bibinfo{year}{2022}).
\newblock


\bibitem[Hore and Ziou(2010)]%
        {psnr}
\bibfield{author}{\bibinfo{person}{Alain Hore} {and} \bibinfo{person}{Djemel Ziou}.} \bibinfo{year}{2010}\natexlab{}.
\newblock \showarticletitle{Image quality metrics: PSNR vs. SSIM}. In \bibinfo{booktitle}{\emph{2010 20th international conference on pattern recognition}}. IEEE, \bibinfo{pages}{2366--2369}.
\newblock


\bibitem[Hu et~al\mbox{.}(2023)]%
        {hu2023sherf}
\bibfield{author}{\bibinfo{person}{Shoukang Hu}, \bibinfo{person}{Fangzhou Hong}, \bibinfo{person}{Liang Pan}, \bibinfo{person}{Haiyi Mei}, \bibinfo{person}{Lei Yang}, {and} \bibinfo{person}{Ziwei Liu}.} \bibinfo{year}{2023}\natexlab{}.
\newblock \showarticletitle{Sherf: Generalizable human nerf from a single image}. In \bibinfo{booktitle}{\emph{Proceedings of the IEEE/CVF International Conference on Computer Vision}}. \bibinfo{pages}{9352--9364}.
\newblock


\bibitem[Hu et~al\mbox{.}(2024a)]%
        {hu2024gauhuman}
\bibfield{author}{\bibinfo{person}{Shoukang Hu}, \bibinfo{person}{Tao Hu}, {and} \bibinfo{person}{Ziwei Liu}.} \bibinfo{year}{2024}\natexlab{a}.
\newblock \showarticletitle{Gauhuman: Articulated gaussian splatting from monocular human videos}. In \bibinfo{booktitle}{\emph{Proceedings of the IEEE/CVF conference on computer vision and pattern recognition}}. \bibinfo{pages}{20418--20431}.
\newblock


\bibitem[Hu et~al\mbox{.}(2025)]%
        {hu2025ex4dextremeviewpoint4d}
\bibfield{author}{\bibinfo{person}{Tao Hu}, \bibinfo{person}{Haoyang Peng}, \bibinfo{person}{Xiao Liu}, {and} \bibinfo{person}{Yuewen Ma}.} \bibinfo{year}{2025}\natexlab{}.
\newblock \bibinfo{title}{EX-4D: EXtreme Viewpoint 4D Video Synthesis via Depth Watertight Mesh}.
\newblock
\showeprint[arxiv]{2506.05554}~[cs.CV]
\urldef\tempurl%
\url{https://arxiv.org/abs/2506.05554}
\showURL{%
\tempurl}


\bibitem[Hu et~al\mbox{.}(2024b)]%
        {hu2024eva}
\bibfield{author}{\bibinfo{person}{Yingdong Hu}, \bibinfo{person}{Zhening Liu}, \bibinfo{person}{Jiawei Shao}, \bibinfo{person}{Zehong Lin}, {and} \bibinfo{person}{Jun Zhang}.} \bibinfo{year}{2024}\natexlab{b}.
\newblock \showarticletitle{Eva-Gaussian: 3D Gaussian-based real-time human novel view synthesis under diverse camera settings}.
\newblock \bibinfo{journal}{\emph{arXiv preprint arXiv:2410.01425}} (\bibinfo{year}{2024}).
\newblock


\bibitem[Huang et~al\mbox{.}(2024b)]%
        {Huang2DGS2024}
\bibfield{author}{\bibinfo{person}{Binbin Huang}, \bibinfo{person}{Zehao Yu}, \bibinfo{person}{Anpei Chen}, \bibinfo{person}{Andreas Geiger}, {and} \bibinfo{person}{Shenghua Gao}.} \bibinfo{year}{2024}\natexlab{b}.
\newblock \showarticletitle{2D Gaussian Splatting for Geometrically Accurate Radiance Fields}. In \bibinfo{booktitle}{\emph{SIGGRAPH 2024 Conference Papers}}. \bibinfo{publisher}{Association for Computing Machinery}.
\newblock
\href{https://doi.org/10.1145/3641519.3657428}{doi:\nolinkurl{10.1145/3641519.3657428}}


\bibitem[Huang et~al\mbox{.}(2025)]%
        {huang2025mvtokenflow}
\bibfield{author}{\bibinfo{person}{Hanzhuo Huang}, \bibinfo{person}{Yuan Liu}, \bibinfo{person}{Ge Zheng}, \bibinfo{person}{Jiepeng Wang}, \bibinfo{person}{Zhiyang Dou}, {and} \bibinfo{person}{Sibei Yang}.} \bibinfo{year}{2025}\natexlab{}.
\newblock \showarticletitle{Mvtokenflow: High-quality 4d content generation using multiview token flow}.
\newblock \bibinfo{journal}{\emph{arXiv preprint arXiv:2502.11697}} (\bibinfo{year}{2025}).
\newblock


\bibitem[Huang et~al\mbox{.}(2024a)]%
        {huang2024sc}
\bibfield{author}{\bibinfo{person}{Yi-Hua Huang}, \bibinfo{person}{Yang-Tian Sun}, \bibinfo{person}{Ziyi Yang}, \bibinfo{person}{Xiaoyang Lyu}, \bibinfo{person}{Yan-Pei Cao}, {and} \bibinfo{person}{Xiaojuan Qi}.} \bibinfo{year}{2024}\natexlab{a}.
\newblock \showarticletitle{Sc-gs: Sparse-controlled gaussian splatting for editable dynamic scenes}. In \bibinfo{booktitle}{\emph{Proceedings of the IEEE/CVF conference on computer vision and pattern recognition}}. \bibinfo{pages}{4220--4230}.
\newblock


\bibitem[Huang et~al\mbox{.}(2018)]%
        {huang2018deep}
\bibfield{author}{\bibinfo{person}{Zeng Huang}, \bibinfo{person}{Tianye Li}, \bibinfo{person}{Weikai Chen}, \bibinfo{person}{Yajie Zhao}, \bibinfo{person}{Jun Xing}, \bibinfo{person}{Chloe LeGendre}, \bibinfo{person}{Linjie Luo}, \bibinfo{person}{Chongyang Ma}, {and} \bibinfo{person}{Hao Li}.} \bibinfo{year}{2018}\natexlab{}.
\newblock \showarticletitle{Deep volumetric video from very sparse multi-view performance capture}. In \bibinfo{booktitle}{\emph{Proceedings of the European Conference on Computer Vision (ECCV)}}. \bibinfo{pages}{336--354}.
\newblock


\bibitem[Jiang et~al\mbox{.}(2022a)]%
        {jiang2022selfrecon}
\bibfield{author}{\bibinfo{person}{Boyi Jiang}, \bibinfo{person}{Yang Hong}, \bibinfo{person}{Hujun Bao}, {and} \bibinfo{person}{Juyong Zhang}.} \bibinfo{year}{2022}\natexlab{a}.
\newblock \showarticletitle{Selfrecon: Self reconstruction your digital avatar from monocular video}. In \bibinfo{booktitle}{\emph{Proceedings of the IEEE/CVF Conference on Computer Vision and Pattern Recognition}}. \bibinfo{pages}{5605--5615}.
\newblock


\bibitem[Jiang et~al\mbox{.}(2023a)]%
        {jiang2023instantavatar}
\bibfield{author}{\bibinfo{person}{Tianjian Jiang}, \bibinfo{person}{Xu Chen}, \bibinfo{person}{Jie Song}, {and} \bibinfo{person}{Otmar Hilliges}.} \bibinfo{year}{2023}\natexlab{a}.
\newblock \showarticletitle{Instantavatar: Learning avatars from monocular video in 60 seconds}. In \bibinfo{booktitle}{\emph{Proceedings of the IEEE/CVF Conference on Computer Vision and Pattern Recognition}}. \bibinfo{pages}{16922--16932}.
\newblock


\bibitem[Jiang et~al\mbox{.}(2022b)]%
        {jiang2022neuman}
\bibfield{author}{\bibinfo{person}{Wei Jiang}, \bibinfo{person}{Kwang~Moo Yi}, \bibinfo{person}{Golnoosh Samei}, \bibinfo{person}{Oncel Tuzel}, {and} \bibinfo{person}{Anurag Ranjan}.} \bibinfo{year}{2022}\natexlab{b}.
\newblock \showarticletitle{Neuman: Neural human radiance field from a single video}. In \bibinfo{booktitle}{\emph{European Conference on Computer Vision}}. Springer, \bibinfo{pages}{402--418}.
\newblock


\bibitem[Jiang et~al\mbox{.}(2020)]%
        {jiang2020sdfdiff}
\bibfield{author}{\bibinfo{person}{Yue Jiang}, \bibinfo{person}{Dantong Ji}, \bibinfo{person}{Zhizhong Han}, {and} \bibinfo{person}{Matthias Zwicker}.} \bibinfo{year}{2020}\natexlab{}.
\newblock \showarticletitle{Sdfdiff: Differentiable rendering of signed distance fields for 3d shape optimization}. In \bibinfo{booktitle}{\emph{Proceedings of the IEEE/CVF conference on computer vision and pattern recognition}}. \bibinfo{pages}{1251--1261}.
\newblock


\bibitem[Jiang et~al\mbox{.}(2024a)]%
        {jiang2024robust}
\bibfield{author}{\bibinfo{person}{Yuheng Jiang}, \bibinfo{person}{Zhehao Shen}, \bibinfo{person}{Yu Hong}, \bibinfo{person}{Chengcheng Guo}, \bibinfo{person}{Yize Wu}, \bibinfo{person}{Yingliang Zhang}, \bibinfo{person}{Jingyi Yu}, {and} \bibinfo{person}{Lan Xu}.} \bibinfo{year}{2024}\natexlab{a}.
\newblock \showarticletitle{Robust dual gaussian splatting for immersive human-centric volumetric videos}.
\newblock \bibinfo{journal}{\emph{ACM Transactions on Graphics (TOG)}} \bibinfo{volume}{43}, \bibinfo{number}{6} (\bibinfo{year}{2024}), \bibinfo{pages}{1--15}.
\newblock


\bibitem[Jiang et~al\mbox{.}(2024b)]%
        {jiang2024hifi4g}
\bibfield{author}{\bibinfo{person}{Yuheng Jiang}, \bibinfo{person}{Zhehao Shen}, \bibinfo{person}{Penghao Wang}, \bibinfo{person}{Zhuo Su}, \bibinfo{person}{Yu Hong}, \bibinfo{person}{Yingliang Zhang}, \bibinfo{person}{Jingyi Yu}, {and} \bibinfo{person}{Lan Xu}.} \bibinfo{year}{2024}\natexlab{b}.
\newblock \showarticletitle{Hifi4g: High-fidelity human performance rendering via compact gaussian splatting}. In \bibinfo{booktitle}{\emph{Proceedings of the IEEE/CVF conference on computer vision and pattern recognition}}. \bibinfo{pages}{19734--19745}.
\newblock


\bibitem[Jiang et~al\mbox{.}(2024c)]%
        {jiang2024animate3d}
\bibfield{author}{\bibinfo{person}{Yanqin Jiang}, \bibinfo{person}{Chaohui Yu}, \bibinfo{person}{Chenjie Cao}, \bibinfo{person}{Fan Wang}, \bibinfo{person}{Weiming Hu}, {and} \bibinfo{person}{Jin Gao}.} \bibinfo{year}{2024}\natexlab{c}.
\newblock \showarticletitle{Animate3d: Animating any 3d model with multi-view video diffusion}.
\newblock \bibinfo{journal}{\emph{arXiv preprint arXiv:2407.11398}} (\bibinfo{year}{2024}).
\newblock


\bibitem[Jiang et~al\mbox{.}(2023b)]%
        {jiang2023consistent4d}
\bibfield{author}{\bibinfo{person}{Yanqin Jiang}, \bibinfo{person}{Li Zhang}, \bibinfo{person}{Jin Gao}, \bibinfo{person}{Weimin Hu}, {and} \bibinfo{person}{Yao Yao}.} \bibinfo{year}{2023}\natexlab{b}.
\newblock \showarticletitle{Consistent4d: Consistent 360 $\{$$\backslash$deg$\}$ dynamic object generation from monocular video}.
\newblock \bibinfo{journal}{\emph{arXiv preprint arXiv:2311.02848}} (\bibinfo{year}{2023}).
\newblock


\bibitem[Jin et~al\mbox{.}(2025)]%
        {jin2025diffuman4d}
\bibfield{author}{\bibinfo{person}{Yudong Jin}, \bibinfo{person}{Sida Peng}, \bibinfo{person}{Xuan Wang}, \bibinfo{person}{Tao Xie}, \bibinfo{person}{Zhen Xu}, \bibinfo{person}{Yifan Yang}, \bibinfo{person}{Yujun Shen}, \bibinfo{person}{Hujun Bao}, {and} \bibinfo{person}{Xiaowei Zhou}.} \bibinfo{year}{2025}\natexlab{}.
\newblock \showarticletitle{Diffuman4D: 4D Consistent Human View Synthesis from Sparse-View Videos with Spatio-Temporal Diffusion Models}. In \bibinfo{booktitle}{\emph{International Conference on Computer Vision (ICCV)}}.
\newblock


\bibitem[Kant et~al\mbox{.}(2025)]%
        {Kant2024Pippo}
\bibfield{author}{\bibinfo{person}{Yash Kant}, \bibinfo{person}{Ethan Weber}, \bibinfo{person}{Jin~Kyu Kim}, \bibinfo{person}{Rawal Khirodkar}, \bibinfo{person}{Su Zhaoen}, \bibinfo{person}{Julieta Martinez}, \bibinfo{person}{Igor Gilitschenski}, \bibinfo{person}{Shunsuke Saito}, {and} \bibinfo{person}{Timur Bagautdinov}.} \bibinfo{year}{2025}\natexlab{}.
\newblock \showarticletitle{Pippo: High-Resolution Multi-View Humans from a Single Image}.
\newblock  (\bibinfo{year}{2025}).
\newblock


\bibitem[Kerbl et~al\mbox{.}(2023)]%
        {kerbl20233d}
\bibfield{author}{\bibinfo{person}{Bernhard Kerbl}, \bibinfo{person}{Georgios Kopanas}, \bibinfo{person}{Thomas Leimk{\"u}hler}, {and} \bibinfo{person}{George Drettakis}.} \bibinfo{year}{2023}\natexlab{}.
\newblock \showarticletitle{3d gaussian splatting for real-time radiance field rendering.}
\newblock \bibinfo{journal}{\emph{ACM Trans. Graph.}} \bibinfo{volume}{42}, \bibinfo{number}{4} (\bibinfo{year}{2023}), \bibinfo{pages}{139--1}.
\newblock


\bibitem[Khirodkar et~al\mbox{.}(2024)]%
        {khirodkar2024sapiens}
\bibfield{author}{\bibinfo{person}{Rawal Khirodkar}, \bibinfo{person}{Timur Bagautdinov}, \bibinfo{person}{Julieta Martinez}, \bibinfo{person}{Su Zhaoen}, \bibinfo{person}{Austin James}, \bibinfo{person}{Peter Selednik}, \bibinfo{person}{Stuart Anderson}, {and} \bibinfo{person}{Shunsuke Saito}.} \bibinfo{year}{2024}\natexlab{}.
\newblock \showarticletitle{Sapiens: Foundation for Human Vision Models}.
\newblock \bibinfo{journal}{\emph{arXiv preprint arXiv:2408.12569}} (\bibinfo{year}{2024}).
\newblock


\bibitem[Kocabas et~al\mbox{.}(2024)]%
        {kocabas2024hugs}
\bibfield{author}{\bibinfo{person}{Muhammed Kocabas}, \bibinfo{person}{Jen-Hao~Rick Chang}, \bibinfo{person}{James Gabriel}, \bibinfo{person}{Oncel Tuzel}, {and} \bibinfo{person}{Anurag Ranjan}.} \bibinfo{year}{2024}\natexlab{}.
\newblock \showarticletitle{{HUGS}: Human Gaussian Splatting}. In \bibinfo{booktitle}{\emph{2024 IEEE/CVF Conference on Computer Vision and Pattern Recognition (CVPR)}}.
\newblock
\urldef\tempurl%
\url{https://arxiv.org/abs/2311.17910}
\showURL{%
\tempurl}


\bibitem[Kwon et~al\mbox{.}(2024)]%
        {kwon2024generalizable}
\bibfield{author}{\bibinfo{person}{Youngjoong Kwon}, \bibinfo{person}{Baole Fang}, \bibinfo{person}{Yixing Lu}, \bibinfo{person}{Haoye Dong}, \bibinfo{person}{Cheng Zhang}, \bibinfo{person}{Francisco~Vicente Carrasco}, \bibinfo{person}{Albert Mosella-Montoro}, \bibinfo{person}{Jianjin Xu}, \bibinfo{person}{Shingo Takagi}, \bibinfo{person}{Daeil Kim}, {et~al\mbox{.}}} \bibinfo{year}{2024}\natexlab{}.
\newblock \showarticletitle{Generalizable human gaussians for sparse view synthesis}. In \bibinfo{booktitle}{\emph{European Conference on Computer Vision}}. Springer, \bibinfo{pages}{451--468}.
\newblock


\bibitem[Kwon et~al\mbox{.}(2021)]%
        {kwon2021neural}
\bibfield{author}{\bibinfo{person}{Youngjoong Kwon}, \bibinfo{person}{Dahun Kim}, \bibinfo{person}{Duygu Ceylan}, {and} \bibinfo{person}{Henry Fuchs}.} \bibinfo{year}{2021}\natexlab{}.
\newblock \showarticletitle{Neural human performer: Learning generalizable radiance fields for human performance rendering}.
\newblock \bibinfo{journal}{\emph{Advances in Neural Information Processing Systems}}  \bibinfo{volume}{34} (\bibinfo{year}{2021}), \bibinfo{pages}{24741--24752}.
\newblock


\bibitem[Levoy and Hanrahan(2023)]%
        {levoy2023light}
\bibfield{author}{\bibinfo{person}{Marc Levoy} {and} \bibinfo{person}{Pat Hanrahan}.} \bibinfo{year}{2023}\natexlab{}.
\newblock \showarticletitle{Light field rendering}.
\newblock In \bibinfo{booktitle}{\emph{Seminal Graphics Papers: Pushing the Boundaries, Volume 2}}. \bibinfo{pages}{441--452}.
\newblock


\bibitem[Li et~al\mbox{.}(2025)]%
        {li2025mvhumannet++}
\bibfield{author}{\bibinfo{person}{Chenghong Li}, \bibinfo{person}{Hongjie Liao}, \bibinfo{person}{Yihao Zhi}, \bibinfo{person}{Xihe Yang}, \bibinfo{person}{Zhengwentai Sun}, \bibinfo{person}{Jiahao Chang}, \bibinfo{person}{Shuguang Cui}, {and} \bibinfo{person}{Xiaoguang Han}.} \bibinfo{year}{2025}\natexlab{}.
\newblock \showarticletitle{MVHumanNet++: A Large-scale Dataset of Multi-view Daily Dressing Human Captures with Richer Annotations for 3D Human Digitization}.
\newblock \bibinfo{journal}{\emph{arXiv preprint arXiv:2505.01838}} (\bibinfo{year}{2025}).
\newblock


\bibitem[Li et~al\mbox{.}(2022)]%
        {li2022tava}
\bibfield{author}{\bibinfo{person}{Ruilong Li}, \bibinfo{person}{Julian Tanke}, \bibinfo{person}{Minh Vo}, \bibinfo{person}{Michael Zollh{\"o}fer}, \bibinfo{person}{J{\"u}rgen Gall}, \bibinfo{person}{Angjoo Kanazawa}, {and} \bibinfo{person}{Christoph Lassner}.} \bibinfo{year}{2022}\natexlab{}.
\newblock \showarticletitle{Tava: Template-free animatable volumetric actors}. In \bibinfo{booktitle}{\emph{European Conference on Computer Vision}}. Springer, \bibinfo{pages}{419--436}.
\newblock


\bibitem[Li et~al\mbox{.}(2024a)]%
        {li2024spacetime}
\bibfield{author}{\bibinfo{person}{Zhan Li}, \bibinfo{person}{Zhang Chen}, \bibinfo{person}{Zhong Li}, {and} \bibinfo{person}{Yi Xu}.} \bibinfo{year}{2024}\natexlab{a}.
\newblock \showarticletitle{Spacetime gaussian feature splatting for real-time dynamic view synthesis}. In \bibinfo{booktitle}{\emph{Proceedings of the IEEE/CVF Conference on Computer Vision and Pattern Recognition}}. \bibinfo{pages}{8508--8520}.
\newblock


\bibitem[Li et~al\mbox{.}(2024b)]%
        {li2024_megasam}
\bibfield{author}{\bibinfo{person}{Zhengqi Li}, \bibinfo{person}{Richard Tucker}, \bibinfo{person}{Forrester Cole}, \bibinfo{person}{Qianqian Wang}, \bibinfo{person}{Linyi Jin}, \bibinfo{person}{Vickie Ye}, \bibinfo{person}{Angjoo Kanazawa}, \bibinfo{person}{Aleksander Holynski}, {and} \bibinfo{person}{Noah Snavely}.} \bibinfo{year}{2024}\natexlab{b}.
\newblock \showarticletitle{MegaSaM: Accurate, Fast and Robust Structure and Motion from Casual Dynamic Videos}. In \bibinfo{booktitle}{\emph{arxiv}}.
\newblock


\bibitem[Li et~al\mbox{.}(2023)]%
        {li2023posevocab}
\bibfield{author}{\bibinfo{person}{Zhe Li}, \bibinfo{person}{Zerong Zheng}, \bibinfo{person}{Yuxiao Liu}, \bibinfo{person}{Boyao Zhou}, {and} \bibinfo{person}{Yebin Liu}.} \bibinfo{year}{2023}\natexlab{}.
\newblock \showarticletitle{Posevocab: Learning joint-structured pose embeddings for human avatar modeling}. In \bibinfo{booktitle}{\emph{ACM SIGGRAPH 2023 conference proceedings}}. \bibinfo{pages}{1--11}.
\newblock


\bibitem[Li et~al\mbox{.}(2024c)]%
        {li2024animatable}
\bibfield{author}{\bibinfo{person}{Zhe Li}, \bibinfo{person}{Zerong Zheng}, \bibinfo{person}{Lizhen Wang}, {and} \bibinfo{person}{Yebin Liu}.} \bibinfo{year}{2024}\natexlab{c}.
\newblock \showarticletitle{Animatable gaussians: Learning pose-dependent gaussian maps for high-fidelity human avatar modeling}. In \bibinfo{booktitle}{\emph{Proceedings of the IEEE/CVF conference on computer vision and pattern recognition}}. \bibinfo{pages}{19711--19722}.
\newblock


\bibitem[Lin et~al\mbox{.}(2024b)]%
        {lin2024open}
\bibfield{author}{\bibinfo{person}{Bin Lin}, \bibinfo{person}{Yunyang Ge}, \bibinfo{person}{Xinhua Cheng}, \bibinfo{person}{Zongjian Li}, \bibinfo{person}{Bin Zhu}, \bibinfo{person}{Shaodong Wang}, \bibinfo{person}{Xianyi He}, \bibinfo{person}{Yang Ye}, \bibinfo{person}{Shenghai Yuan}, \bibinfo{person}{Liuhan Chen}, {et~al\mbox{.}}} \bibinfo{year}{2024}\natexlab{b}.
\newblock \showarticletitle{Open-sora plan: Open-source large video generation model}.
\newblock \bibinfo{journal}{\emph{arXiv preprint arXiv:2412.00131}} (\bibinfo{year}{2024}).
\newblock


\bibitem[Lin et~al\mbox{.}(2024a)]%
        {lin2024gaussian}
\bibfield{author}{\bibinfo{person}{Youtian Lin}, \bibinfo{person}{Zuozhuo Dai}, \bibinfo{person}{Siyu Zhu}, {and} \bibinfo{person}{Yao Yao}.} \bibinfo{year}{2024}\natexlab{a}.
\newblock \showarticletitle{Gaussian-flow: 4d reconstruction with dynamic 3d gaussian particle}. In \bibinfo{booktitle}{\emph{Proceedings of the IEEE/CVF Conference on Computer Vision and Pattern Recognition}}. \bibinfo{pages}{21136--21145}.
\newblock


\bibitem[Ling et~al\mbox{.}(2024)]%
        {ling2024align}
\bibfield{author}{\bibinfo{person}{Huan Ling}, \bibinfo{person}{Seung~Wook Kim}, \bibinfo{person}{Antonio Torralba}, \bibinfo{person}{Sanja Fidler}, {and} \bibinfo{person}{Karsten Kreis}.} \bibinfo{year}{2024}\natexlab{}.
\newblock \showarticletitle{Align your gaussians: Text-to-4d with dynamic 3d gaussians and composed diffusion models}. In \bibinfo{booktitle}{\emph{Proceedings of the IEEE/CVF conference on computer vision and pattern recognition}}. \bibinfo{pages}{8576--8588}.
\newblock


\bibitem[Lipman et~al\mbox{.}(2022)]%
        {lipman2022flow}
\bibfield{author}{\bibinfo{person}{Yaron Lipman}, \bibinfo{person}{Ricky~TQ Chen}, \bibinfo{person}{Heli Ben-Hamu}, \bibinfo{person}{Maximilian Nickel}, {and} \bibinfo{person}{Matt Le}.} \bibinfo{year}{2022}\natexlab{}.
\newblock \showarticletitle{Flow matching for generative modeling}.
\newblock \bibinfo{journal}{\emph{arXiv preprint arXiv:2210.02747}} (\bibinfo{year}{2022}).
\newblock


\bibitem[Liu et~al\mbox{.}(2021)]%
        {liu2021neural}
\bibfield{author}{\bibinfo{person}{Lingjie Liu}, \bibinfo{person}{Marc Habermann}, \bibinfo{person}{Viktor Rudnev}, \bibinfo{person}{Kripasindhu Sarkar}, \bibinfo{person}{Jiatao Gu}, {and} \bibinfo{person}{Christian Theobalt}.} \bibinfo{year}{2021}\natexlab{}.
\newblock \showarticletitle{Neural actor: Neural free-view synthesis of human actors with pose control}.
\newblock \bibinfo{journal}{\emph{ACM transactions on graphics (TOG)}} \bibinfo{volume}{40}, \bibinfo{number}{6} (\bibinfo{year}{2021}), \bibinfo{pages}{1--16}.
\newblock


\bibitem[Liu et~al\mbox{.}(2024)]%
        {liu2024one}
\bibfield{author}{\bibinfo{person}{Minghua Liu}, \bibinfo{person}{Ruoxi Shi}, \bibinfo{person}{Linghao Chen}, \bibinfo{person}{Zhuoyang Zhang}, \bibinfo{person}{Chao Xu}, \bibinfo{person}{Xinyue Wei}, \bibinfo{person}{Hansheng Chen}, \bibinfo{person}{Chong Zeng}, \bibinfo{person}{Jiayuan Gu}, {and} \bibinfo{person}{Hao Su}.} \bibinfo{year}{2024}\natexlab{}.
\newblock \showarticletitle{One-2-3-45++: Fast single image to 3d objects with consistent multi-view generation and 3d diffusion}. In \bibinfo{booktitle}{\emph{Proceedings of the IEEE/CVF conference on computer vision and pattern recognition}}. \bibinfo{pages}{10072--10083}.
\newblock


\bibitem[Liu et~al\mbox{.}(2023c)]%
        {liu2023one}
\bibfield{author}{\bibinfo{person}{Minghua Liu}, \bibinfo{person}{Chao Xu}, \bibinfo{person}{Haian Jin}, \bibinfo{person}{Linghao Chen}, \bibinfo{person}{Mukund Varma~T}, \bibinfo{person}{Zexiang Xu}, {and} \bibinfo{person}{Hao Su}.} \bibinfo{year}{2023}\natexlab{c}.
\newblock \showarticletitle{One-2-3-45: Any single image to 3d mesh in 45 seconds without per-shape optimization}.
\newblock \bibinfo{journal}{\emph{Advances in Neural Information Processing Systems}}  \bibinfo{volume}{36} (\bibinfo{year}{2023}), \bibinfo{pages}{22226--22246}.
\newblock


\bibitem[Liu et~al\mbox{.}(2023b)]%
        {liu2023zero}
\bibfield{author}{\bibinfo{person}{Ruoshi Liu}, \bibinfo{person}{Rundi Wu}, \bibinfo{person}{Basile Van~Hoorick}, \bibinfo{person}{Pavel Tokmakov}, \bibinfo{person}{Sergey Zakharov}, {and} \bibinfo{person}{Carl Vondrick}.} \bibinfo{year}{2023}\natexlab{b}.
\newblock \showarticletitle{Zero-1-to-3: Zero-shot one image to 3d object}. In \bibinfo{booktitle}{\emph{Proceedings of the IEEE/CVF international conference on computer vision}}. \bibinfo{pages}{9298--9309}.
\newblock


\bibitem[Liu et~al\mbox{.}(2025)]%
        {liu2025free4d}
\bibfield{author}{\bibinfo{person}{Tianqi Liu}, \bibinfo{person}{Zihao Huang}, \bibinfo{person}{Zhaoxi Chen}, \bibinfo{person}{Guangcong Wang}, \bibinfo{person}{Shoukang Hu}, \bibinfo{person}{liao Shen}, \bibinfo{person}{Huiqiang Sun}, \bibinfo{person}{Zhiguo Cao}, \bibinfo{person}{Wei Li}, {and} \bibinfo{person}{Ziwei Liu}.} \bibinfo{year}{2025}\natexlab{}.
\newblock \showarticletitle{Free4D: Tuning-free 4D Scene Generation with Spatial-Temporal Consistency}.
\newblock \bibinfo{journal}{\emph{arXiv preprint arXiv:2503.20785}} (\bibinfo{year}{2025}).
\newblock


\bibitem[Liu et~al\mbox{.}(2023a)]%
        {liu2023syncdreamer}
\bibfield{author}{\bibinfo{person}{Yuan Liu}, \bibinfo{person}{Cheng Lin}, \bibinfo{person}{Zijiao Zeng}, \bibinfo{person}{Xiaoxiao Long}, \bibinfo{person}{Lingjie Liu}, \bibinfo{person}{Taku Komura}, {and} \bibinfo{person}{Wenping Wang}.} \bibinfo{year}{2023}\natexlab{a}.
\newblock \showarticletitle{Syncdreamer: Generating multiview-consistent images from a single-view image}.
\newblock \bibinfo{journal}{\emph{arXiv preprint arXiv:2309.03453}} (\bibinfo{year}{2023}).
\newblock


\bibitem[Loper et~al\mbox{.}(2015)]%
        {loper2015smpl}
\bibfield{author}{\bibinfo{person}{Matthew Loper}, \bibinfo{person}{Naureen Mahmood}, \bibinfo{person}{Javier Romero}, \bibinfo{person}{Gerard Pons-Moll}, {and} \bibinfo{person}{Michael~J. Black}.} \bibinfo{year}{2015}\natexlab{}.
\newblock \showarticletitle{{SMPL}: A Skinned Multi-Person Linear Model}.
\newblock \bibinfo{journal}{\emph{ACM Trans. Graphics (Proc. SIGGRAPH Asia)}} \bibinfo{volume}{34}, \bibinfo{number}{6} (\bibinfo{date}{Oct.} \bibinfo{year}{2015}), \bibinfo{pages}{248:1--248:16}.
\newblock


\bibitem[Loshchilov and Hutter(2017)]%
        {loshchilov2017decoupled}
\bibfield{author}{\bibinfo{person}{Ilya Loshchilov} {and} \bibinfo{person}{Frank Hutter}.} \bibinfo{year}{2017}\natexlab{}.
\newblock \showarticletitle{Decoupled weight decay regularization}.
\newblock \bibinfo{journal}{\emph{arXiv preprint arXiv:1711.05101}} (\bibinfo{year}{2017}).
\newblock


\bibitem[Lu et~al\mbox{.}(2025)]%
        {lu2025gas}
\bibfield{author}{\bibinfo{person}{Yixing Lu}, \bibinfo{person}{Junting Dong}, \bibinfo{person}{Youngjoong Kwon}, \bibinfo{person}{Qin Zhao}, \bibinfo{person}{Bo Dai}, {and} \bibinfo{person}{Fernando De~la Torre}.} \bibinfo{year}{2025}\natexlab{}.
\newblock \showarticletitle{GAS: Generative Avatar Synthesis from a Single Image}.
\newblock \bibinfo{journal}{\emph{arXiv preprint arXiv:2502.06957}} (\bibinfo{year}{2025}).
\newblock


\bibitem[Mihajlovic et~al\mbox{.}(2022)]%
        {mihajlovic2022keypointnerf}
\bibfield{author}{\bibinfo{person}{Marko Mihajlovic}, \bibinfo{person}{Aayush Bansal}, \bibinfo{person}{Michael Zollhoefer}, \bibinfo{person}{Siyu Tang}, {and} \bibinfo{person}{Shunsuke Saito}.} \bibinfo{year}{2022}\natexlab{}.
\newblock \showarticletitle{KeypointNeRF: Generalizing image-based volumetric avatars using relative spatial encoding of keypoints}. In \bibinfo{booktitle}{\emph{European conference on computer vision}}. Springer, \bibinfo{pages}{179--197}.
\newblock


\bibitem[Mildenhall et~al\mbox{.}(2020)]%
        {mildenhall2020nerf}
\bibfield{author}{\bibinfo{person}{Ben Mildenhall}, \bibinfo{person}{Pratul~P. Srinivasan}, \bibinfo{person}{Matthew Tancik}, \bibinfo{person}{Jonathan~T. Barron}, \bibinfo{person}{Ravi Ramamoorthi}, {and} \bibinfo{person}{Ren Ng}.} \bibinfo{year}{2020}\natexlab{}.
\newblock \showarticletitle{NeRF: Representing Scenes as Neural Radiance Fields for View Synthesis}. In \bibinfo{booktitle}{\emph{ECCV}}.
\newblock


\bibitem[Mildenhall et~al\mbox{.}(2021)]%
        {nerf}
\bibfield{author}{\bibinfo{person}{Ben Mildenhall}, \bibinfo{person}{Pratul~P Srinivasan}, \bibinfo{person}{Matthew Tancik}, \bibinfo{person}{Jonathan~T Barron}, \bibinfo{person}{Ravi Ramamoorthi}, {and} \bibinfo{person}{Ren Ng}.} \bibinfo{year}{2021}\natexlab{}.
\newblock \showarticletitle{Nerf: Representing scenes as neural radiance fields for view synthesis}.
\newblock \bibinfo{journal}{\emph{Commun. ACM}} \bibinfo{volume}{65}, \bibinfo{number}{1} (\bibinfo{year}{2021}), \bibinfo{pages}{99--106}.
\newblock


\bibitem[Orts-Escolano et~al\mbox{.}(2016)]%
        {orts2016holoportation}
\bibfield{author}{\bibinfo{person}{Sergio Orts-Escolano}, \bibinfo{person}{Christoph Rhemann}, \bibinfo{person}{Sean Fanello}, \bibinfo{person}{Wayne Chang}, \bibinfo{person}{Adarsh Kowdle}, \bibinfo{person}{Yury Degtyarev}, \bibinfo{person}{David Kim}, \bibinfo{person}{Philip~L Davidson}, \bibinfo{person}{Sameh Khamis}, \bibinfo{person}{Mingsong Dou}, {et~al\mbox{.}}} \bibinfo{year}{2016}\natexlab{}.
\newblock \showarticletitle{Holoportation: Virtual 3d teleportation in real-time}. In \bibinfo{booktitle}{\emph{Proceedings of the 29th annual symposium on user interface software and technology}}. \bibinfo{pages}{741--754}.
\newblock


\bibitem[Pang et~al\mbox{.}(2024)]%
        {pang2024ash}
\bibfield{author}{\bibinfo{person}{Haokai Pang}, \bibinfo{person}{Heming Zhu}, \bibinfo{person}{Adam Kortylewski}, \bibinfo{person}{Christian Theobalt}, {and} \bibinfo{person}{Marc Habermann}.} \bibinfo{year}{2024}\natexlab{}.
\newblock \showarticletitle{Ash: Animatable gaussian splats for efficient and photoreal human rendering}. In \bibinfo{booktitle}{\emph{Proceedings of the IEEE/CVF Conference on Computer Vision and Pattern Recognition}}. \bibinfo{pages}{1165--1175}.
\newblock


\bibitem[Pang et~al\mbox{.}(2025)]%
        {Disco4D}
\bibfield{author}{\bibinfo{person}{Hui~En Pang}, \bibinfo{person}{Shuai Liu}, \bibinfo{person}{Zhongang Cai}, \bibinfo{person}{Lei Yang}, \bibinfo{person}{Tianwei Zhang}, {and} \bibinfo{person}{Ziwei Liu}.} \bibinfo{year}{2025}\natexlab{}.
\newblock \showarticletitle{Disco4D: Disentangled 4D Human Generation and Animation from a Single Image}. In \bibinfo{booktitle}{\emph{CVPR}}.
\newblock


\bibitem[Park et~al\mbox{.}(2019)]%
        {park2019deepsdf}
\bibfield{author}{\bibinfo{person}{Jeong~Joon Park}, \bibinfo{person}{Peter Florence}, \bibinfo{person}{Julian Straub}, \bibinfo{person}{Richard Newcombe}, {and} \bibinfo{person}{Steven Lovegrove}.} \bibinfo{year}{2019}\natexlab{}.
\newblock \showarticletitle{Deepsdf: Learning continuous signed distance functions for shape representation}. In \bibinfo{booktitle}{\emph{Proceedings of the IEEE/CVF conference on computer vision and pattern recognition}}. \bibinfo{pages}{165--174}.
\newblock


\bibitem[Pavlakos et~al\mbox{.}(2019)]%
        {pavlakos2019smplx}
\bibfield{author}{\bibinfo{person}{Georgios Pavlakos}, \bibinfo{person}{Vasileios Choutas}, \bibinfo{person}{Nima Ghorbani}, \bibinfo{person}{Timo Bolkart}, \bibinfo{person}{Ahmed A.~A. Osman}, \bibinfo{person}{Dimitrios Tzionas}, {and} \bibinfo{person}{Michael~J. Black}.} \bibinfo{year}{2019}\natexlab{}.
\newblock \showarticletitle{Expressive Body Capture: {3D} Hands, Face, and Body from a Single Image}. In \bibinfo{booktitle}{\emph{Proceedings IEEE Conf. on Computer Vision and Pattern Recognition (CVPR)}}. \bibinfo{pages}{10975--10985}.
\newblock


\bibitem[Peng et~al\mbox{.}(2021a)]%
        {peng2021animatable}
\bibfield{author}{\bibinfo{person}{Sida Peng}, \bibinfo{person}{Junting Dong}, \bibinfo{person}{Qianqian Wang}, \bibinfo{person}{Shangzhan Zhang}, \bibinfo{person}{Qing Shuai}, \bibinfo{person}{Xiaowei Zhou}, {and} \bibinfo{person}{Hujun Bao}.} \bibinfo{year}{2021}\natexlab{a}.
\newblock \showarticletitle{Animatable neural radiance fields for modeling dynamic human bodies}. In \bibinfo{booktitle}{\emph{Proceedings of the IEEE/CVF International Conference on Computer Vision}}. \bibinfo{pages}{14314--14323}.
\newblock


\bibitem[Peng et~al\mbox{.}(2021b)]%
        {peng2021neural}
\bibfield{author}{\bibinfo{person}{Sida Peng}, \bibinfo{person}{Yuanqing Zhang}, \bibinfo{person}{Yinghao Xu}, \bibinfo{person}{Qianqian Wang}, \bibinfo{person}{Qing Shuai}, \bibinfo{person}{Hujun Bao}, {and} \bibinfo{person}{Xiaowei Zhou}.} \bibinfo{year}{2021}\natexlab{b}.
\newblock \showarticletitle{Neural body: Implicit neural representations with structured latent codes for novel view synthesis of dynamic humans}. In \bibinfo{booktitle}{\emph{Proceedings of the IEEE/CVF conference on computer vision and pattern recognition}}. \bibinfo{pages}{9054--9063}.
\newblock


\bibitem[Piccinelli et~al\mbox{.}(2025)]%
        {piccinelli2025unidepthv2}
\bibfield{author}{\bibinfo{person}{Luigi Piccinelli}, \bibinfo{person}{Christos Sakaridis}, \bibinfo{person}{Yung-Hsu Yang}, \bibinfo{person}{Mattia Segu}, \bibinfo{person}{Siyuan Li}, \bibinfo{person}{Wim Abbeloos}, {and} \bibinfo{person}{Luc~Van Gool}.} \bibinfo{year}{2025}\natexlab{}.
\newblock \bibinfo{title}{{U}ni{D}epth{V2}: Universal Monocular Metric Depth Estimation Made Simpler}.
\newblock
\showeprint[arxiv]{2502.20110}~[cs.CV]
\urldef\tempurl%
\url{https://arxiv.org/abs/2502.20110}
\showURL{%
\tempurl}


\bibitem[Pumarola et~al\mbox{.}(2021)]%
        {pumarola2021d}
\bibfield{author}{\bibinfo{person}{Albert Pumarola}, \bibinfo{person}{Enric Corona}, \bibinfo{person}{Gerard Pons-Moll}, {and} \bibinfo{person}{Francesc Moreno-Noguer}.} \bibinfo{year}{2021}\natexlab{}.
\newblock \showarticletitle{D-nerf: Neural radiance fields for dynamic scenes}. In \bibinfo{booktitle}{\emph{Proceedings of the IEEE/CVF conference on computer vision and pattern recognition}}. \bibinfo{pages}{10318--10327}.
\newblock


\bibitem[Qian et~al\mbox{.}(2024a)]%
        {Qian_2024_CVPR}
\bibfield{author}{\bibinfo{person}{Shenhan Qian}, \bibinfo{person}{Tobias Kirschstein}, \bibinfo{person}{Liam Schoneveld}, \bibinfo{person}{Davide Davoli}, \bibinfo{person}{Simon Giebenhain}, {and} \bibinfo{person}{Matthias Nie{\ss}ner}.} \bibinfo{year}{2024}\natexlab{a}.
\newblock \showarticletitle{GaussianAvatars: Photorealistic Head Avatars with Rigged 3D Gaussians}. In \bibinfo{booktitle}{\emph{Proceedings of the IEEE/CVF Conference on Computer Vision and Pattern Recognition (CVPR)}}. \bibinfo{pages}{20299--20309}.
\newblock


\bibitem[Qian et~al\mbox{.}(2024b)]%
        {qian20233dgsavatar}
\bibfield{author}{\bibinfo{person}{Zhiyin Qian}, \bibinfo{person}{Shaofei Wang}, \bibinfo{person}{Marko Mihajlovic}, \bibinfo{person}{Andreas Geiger}, {and} \bibinfo{person}{Siyu Tang}.} \bibinfo{year}{2024}\natexlab{b}.
\newblock \showarticletitle{3DGS-Avatar: Animatable Avatars via Deformable 3D Gaussian Splatting}. In \bibinfo{booktitle}{\emph{CVPR}}.
\newblock


\bibitem[Ren et~al\mbox{.}(2023)]%
        {ren2023dreamgaussian4d}
\bibfield{author}{\bibinfo{person}{Jiawei Ren}, \bibinfo{person}{Liang Pan}, \bibinfo{person}{Jiaxiang Tang}, \bibinfo{person}{Chi Zhang}, \bibinfo{person}{Ang Cao}, \bibinfo{person}{Gang Zeng}, {and} \bibinfo{person}{Ziwei Liu}.} \bibinfo{year}{2023}\natexlab{}.
\newblock \showarticletitle{DreamGaussian4D: Generative 4D Gaussian Splatting}.
\newblock \bibinfo{journal}{\emph{arXiv preprint arXiv:2312.17142}} (\bibinfo{year}{2023}).
\newblock


\bibitem[Ren et~al\mbox{.}(2025)]%
        {ren2025gen3c}
\bibfield{author}{\bibinfo{person}{Xuanchi Ren}, \bibinfo{person}{Tianchang Shen}, \bibinfo{person}{Jiahui Huang}, \bibinfo{person}{Huan Ling}, \bibinfo{person}{Yifan Lu}, \bibinfo{person}{Merlin Nimier-David}, \bibinfo{person}{Thomas Müller}, \bibinfo{person}{Alexander Keller}, \bibinfo{person}{Sanja Fidler}, {and} \bibinfo{person}{Jun Gao}.} \bibinfo{year}{2025}\natexlab{}.
\newblock \showarticletitle{GEN3C: 3D-Informed World-Consistent Video Generation with Precise Camera Control}. In \bibinfo{booktitle}{\emph{Proceedings of the IEEE/CVF Conference on Computer Vision and Pattern Recognition}}.
\newblock


\bibitem[Rombach et~al\mbox{.}(2022)]%
        {rombach2022high}
\bibfield{author}{\bibinfo{person}{Robin Rombach}, \bibinfo{person}{Andreas Blattmann}, \bibinfo{person}{Dominik Lorenz}, \bibinfo{person}{Patrick Esser}, {and} \bibinfo{person}{Bj{\"o}rn Ommer}.} \bibinfo{year}{2022}\natexlab{}.
\newblock \showarticletitle{High-resolution image synthesis with latent diffusion models}. In \bibinfo{booktitle}{\emph{Proceedings of the IEEE/CVF conference on computer vision and pattern recognition}}. \bibinfo{pages}{10684--10695}.
\newblock


\bibitem[Seitz et~al\mbox{.}(2006)]%
        {seitz2006comparison}
\bibfield{author}{\bibinfo{person}{Steven~M Seitz}, \bibinfo{person}{Brian Curless}, \bibinfo{person}{James Diebel}, \bibinfo{person}{Daniel Scharstein}, {and} \bibinfo{person}{Richard Szeliski}.} \bibinfo{year}{2006}\natexlab{}.
\newblock \showarticletitle{A comparison and evaluation of multi-view stereo reconstruction algorithms}. In \bibinfo{booktitle}{\emph{2006 IEEE computer society conference on computer vision and pattern recognition (CVPR'06)}}, Vol.~\bibinfo{volume}{1}. IEEE, \bibinfo{pages}{519--528}.
\newblock


\bibitem[Shao et~al\mbox{.}(2024)]%
        {shao2024human4dit}
\bibfield{author}{\bibinfo{person}{Ruizhi Shao}, \bibinfo{person}{Youxin Pang}, \bibinfo{person}{Zerong Zheng}, \bibinfo{person}{Jingxiang Sun}, {and} \bibinfo{person}{Yebin Liu}.} \bibinfo{year}{2024}\natexlab{}.
\newblock \showarticletitle{Human4DiT: 360-degree Human Video Generation with 4D Diffusion Transformer}.
\newblock \bibinfo{journal}{\emph{ACM Transactions on Graphics (TOG)}} \bibinfo{volume}{43}, \bibinfo{number}{6} (\bibinfo{year}{2024}).
\newblock


\bibitem[Shao et~al\mbox{.}(2023)]%
        {shao2023tensor4d}
\bibfield{author}{\bibinfo{person}{Ruizhi Shao}, \bibinfo{person}{Zerong Zheng}, \bibinfo{person}{Hanzhang Tu}, \bibinfo{person}{Boning Liu}, \bibinfo{person}{Hongwen Zhang}, {and} \bibinfo{person}{Yebin Liu}.} \bibinfo{year}{2023}\natexlab{}.
\newblock \showarticletitle{Tensor4d: Efficient neural 4d decomposition for high-fidelity dynamic reconstruction and rendering}. In \bibinfo{booktitle}{\emph{Proceedings of the IEEE/CVF Conference on Computer Vision and Pattern Recognition}}. \bibinfo{pages}{16632--16642}.
\newblock


\bibitem[Shen et~al\mbox{.}(2021a)]%
        {shen2021dmtet}
\bibfield{author}{\bibinfo{person}{Tianchang Shen}, \bibinfo{person}{Jun Gao}, \bibinfo{person}{Kangxue Yin}, \bibinfo{person}{Ming-Yu Liu}, {and} \bibinfo{person}{Sanja Fidler}.} \bibinfo{year}{2021}\natexlab{a}.
\newblock \showarticletitle{Deep Marching Tetrahedra: a Hybrid Representation for High-Resolution 3D Shape Synthesis}. In \bibinfo{booktitle}{\emph{Advances in Neural Information Processing Systems (NeurIPS)}}.
\newblock


\bibitem[Shen et~al\mbox{.}(2021b)]%
        {shen2021deep}
\bibfield{author}{\bibinfo{person}{Tianchang Shen}, \bibinfo{person}{Jun Gao}, \bibinfo{person}{Kangxue Yin}, \bibinfo{person}{Ming-Yu Liu}, {and} \bibinfo{person}{Sanja Fidler}.} \bibinfo{year}{2021}\natexlab{b}.
\newblock \showarticletitle{Deep marching tetrahedra: a hybrid representation for high-resolution 3d shape synthesis}.
\newblock \bibinfo{journal}{\emph{Advances in Neural Information Processing Systems}}  \bibinfo{volume}{34} (\bibinfo{year}{2021}), \bibinfo{pages}{6087--6101}.
\newblock


\bibitem[Shi et~al\mbox{.}(2023a)]%
        {shi2023zero123++}
\bibfield{author}{\bibinfo{person}{Ruoxi Shi}, \bibinfo{person}{Hansheng Chen}, \bibinfo{person}{Zhuoyang Zhang}, \bibinfo{person}{Minghua Liu}, \bibinfo{person}{Chao Xu}, \bibinfo{person}{Xinyue Wei}, \bibinfo{person}{Linghao Chen}, \bibinfo{person}{Chong Zeng}, {and} \bibinfo{person}{Hao Su}.} \bibinfo{year}{2023}\natexlab{a}.
\newblock \showarticletitle{Zero123++: a single image to consistent multi-view diffusion base model}.
\newblock \bibinfo{journal}{\emph{arXiv preprint arXiv:2310.15110}} (\bibinfo{year}{2023}).
\newblock


\bibitem[Shi et~al\mbox{.}(2023b)]%
        {shi2023mvdream}
\bibfield{author}{\bibinfo{person}{Yichun Shi}, \bibinfo{person}{Peng Wang}, \bibinfo{person}{Jianglong Ye}, \bibinfo{person}{Mai Long}, \bibinfo{person}{Kejie Li}, {and} \bibinfo{person}{Xiao Yang}.} \bibinfo{year}{2023}\natexlab{b}.
\newblock \showarticletitle{Mvdream: Multi-view diffusion for 3d generation}.
\newblock \bibinfo{journal}{\emph{arXiv preprint arXiv:2308.16512}} (\bibinfo{year}{2023}).
\newblock


\bibitem[Sohl-Dickstein et~al\mbox{.}(2015)]%
        {sohl2015deep}
\bibfield{author}{\bibinfo{person}{Jascha Sohl-Dickstein}, \bibinfo{person}{Eric Weiss}, \bibinfo{person}{Niru Maheswaranathan}, {and} \bibinfo{person}{Surya Ganguli}.} \bibinfo{year}{2015}\natexlab{}.
\newblock \showarticletitle{Deep unsupervised learning using nonequilibrium thermodynamics}. In \bibinfo{booktitle}{\emph{International conference on machine learning}}. pmlr, \bibinfo{pages}{2256--2265}.
\newblock


\bibitem[Song et~al\mbox{.}(2020)]%
        {song2020denoising}
\bibfield{author}{\bibinfo{person}{Jiaming Song}, \bibinfo{person}{Chenlin Meng}, {and} \bibinfo{person}{Stefano Ermon}.} \bibinfo{year}{2020}\natexlab{}.
\newblock \showarticletitle{Denoising diffusion implicit models}.
\newblock \bibinfo{journal}{\emph{arXiv preprint arXiv:2010.02502}} (\bibinfo{year}{2020}).
\newblock


\bibitem[Song and Ermon(2019)]%
        {song2019generative}
\bibfield{author}{\bibinfo{person}{Yang Song} {and} \bibinfo{person}{Stefano Ermon}.} \bibinfo{year}{2019}\natexlab{}.
\newblock \showarticletitle{Generative modeling by estimating gradients of the data distribution}.
\newblock \bibinfo{journal}{\emph{Advances in neural information processing systems}}  \bibinfo{volume}{32} (\bibinfo{year}{2019}).
\newblock


\bibitem[Tang et~al\mbox{.}(2025)]%
        {tang2025gaf}
\bibfield{author}{\bibinfo{person}{Jiapeng Tang}, \bibinfo{person}{Davide Davoli}, \bibinfo{person}{Tobias Kirschstein}, \bibinfo{person}{Liam Schoneveld}, {and} \bibinfo{person}{Matthias Niessner}.} \bibinfo{year}{2025}\natexlab{}.
\newblock \showarticletitle{Gaf: Gaussian avatar reconstruction from monocular videos via multi-view diffusion}. In \bibinfo{booktitle}{\emph{Proceedings of the Computer Vision and Pattern Recognition Conference}}. \bibinfo{pages}{5546--5558}.
\newblock


\bibitem[Tewari et~al\mbox{.}(2020)]%
        {tewari2020state}
\bibfield{author}{\bibinfo{person}{Ayush Tewari}, \bibinfo{person}{Ohad Fried}, \bibinfo{person}{Justus Thies}, \bibinfo{person}{Vincent Sitzmann}, \bibinfo{person}{Stephen Lombardi}, \bibinfo{person}{Kalyan Sunkavalli}, \bibinfo{person}{Ricardo Martin-Brualla}, \bibinfo{person}{Tomas Simon}, \bibinfo{person}{Jason Saragih}, \bibinfo{person}{Matthias Nie{\ss}ner}, {et~al\mbox{.}}} \bibinfo{year}{2020}\natexlab{}.
\newblock \showarticletitle{State of the art on neural rendering}. In \bibinfo{booktitle}{\emph{Computer Graphics Forum}}, Vol.~\bibinfo{volume}{39}. Wiley Online Library, \bibinfo{pages}{701--727}.
\newblock


\bibitem[Thies et~al\mbox{.}(2019)]%
        {thies2019deferred}
\bibfield{author}{\bibinfo{person}{Justus Thies}, \bibinfo{person}{Michael Zollh{\"o}fer}, {and} \bibinfo{person}{Matthias Nie{\ss}ner}.} \bibinfo{year}{2019}\natexlab{}.
\newblock \showarticletitle{Deferred neural rendering: Image synthesis using neural textures}.
\newblock \bibinfo{journal}{\emph{Acm Transactions on Graphics (TOG)}} \bibinfo{volume}{38}, \bibinfo{number}{4} (\bibinfo{year}{2019}), \bibinfo{pages}{1--12}.
\newblock


\bibitem[Unterthiner et~al\mbox{.}(2018)]%
        {unterthiner2018towards}
\bibfield{author}{\bibinfo{person}{Thomas Unterthiner}, \bibinfo{person}{Sjoerd Van~Steenkiste}, \bibinfo{person}{Karol Kurach}, \bibinfo{person}{Raphael Marinier}, \bibinfo{person}{Marcin Michalski}, {and} \bibinfo{person}{Sylvain Gelly}.} \bibinfo{year}{2018}\natexlab{}.
\newblock \showarticletitle{Towards accurate generative models of video: A new metric \& challenges}.
\newblock \bibinfo{journal}{\emph{arXiv preprint arXiv:1812.01717}} (\bibinfo{year}{2018}).
\newblock


\bibitem[Van~Hoorick et~al\mbox{.}(2024)]%
        {vanhoorick2024gcd}
\bibfield{author}{\bibinfo{person}{Basile Van~Hoorick}, \bibinfo{person}{Rundi Wu}, \bibinfo{person}{Ege Ozguroglu}, \bibinfo{person}{Kyle Sargent}, \bibinfo{person}{Ruoshi Liu}, \bibinfo{person}{Pavel Tokmakov}, \bibinfo{person}{Achal Dave}, \bibinfo{person}{Changxi Zheng}, {and} \bibinfo{person}{Carl Vondrick}.} \bibinfo{year}{2024}\natexlab{}.
\newblock \showarticletitle{Generative Camera Dolly: Extreme Monocular Dynamic Novel View Synthesis}.
\newblock \bibinfo{journal}{\emph{European Conference on Computer Vision (ECCV)}} (\bibinfo{year}{2024}).
\newblock


\bibitem[Voleti et~al\mbox{.}(2024)]%
        {voleti2024sv3d}
\bibfield{author}{\bibinfo{person}{Vikram Voleti}, \bibinfo{person}{Chun-Han Yao}, \bibinfo{person}{Mark Boss}, \bibinfo{person}{Adam Letts}, \bibinfo{person}{David Pankratz}, \bibinfo{person}{Dmitry Tochilkin}, \bibinfo{person}{Christian Laforte}, \bibinfo{person}{Robin Rombach}, {and} \bibinfo{person}{Varun Jampani}.} \bibinfo{year}{2024}\natexlab{}.
\newblock \showarticletitle{Sv3d: Novel multi-view synthesis and 3d generation from a single image using latent video diffusion}. In \bibinfo{booktitle}{\emph{European Conference on Computer Vision}}. Springer, \bibinfo{pages}{439--457}.
\newblock


\bibitem[Wan et~al\mbox{.}(2025)]%
        {wan2025wan}
\bibfield{author}{\bibinfo{person}{Team Wan}, \bibinfo{person}{Ang Wang}, \bibinfo{person}{Baole Ai}, \bibinfo{person}{Bin Wen}, \bibinfo{person}{Chaojie Mao}, \bibinfo{person}{Chen-Wei Xie}, \bibinfo{person}{Di Chen}, \bibinfo{person}{Feiwu Yu}, \bibinfo{person}{Haiming Zhao}, \bibinfo{person}{Jianxiao Yang}, {et~al\mbox{.}}} \bibinfo{year}{2025}\natexlab{}.
\newblock \showarticletitle{Wan: Open and advanced large-scale video generative models}.
\newblock \bibinfo{journal}{\emph{arXiv preprint arXiv:2503.20314}} (\bibinfo{year}{2025}).
\newblock


\bibitem[Wang et~al\mbox{.}(2025a)]%
        {wan2025}
\bibfield{author}{\bibinfo{person}{Ang Wang}, \bibinfo{person}{Baole Ai}, \bibinfo{person}{Bin Wen}, \bibinfo{person}{Chaojie Mao}, \bibinfo{person}{Chen-Wei Xie}, \bibinfo{person}{Di Chen}, \bibinfo{person}{Feiwu Yu}, \bibinfo{person}{Haiming Zhao}, \bibinfo{person}{Jianxiao Yang}, \bibinfo{person}{Jianyuan Zeng}, \bibinfo{person}{Jiayu Wang}, \bibinfo{person}{Jingfeng Zhang}, \bibinfo{person}{Jingren Zhou}, \bibinfo{person}{Jinkai Wang}, \bibinfo{person}{Jixuan Chen}, \bibinfo{person}{Kai Zhu}, \bibinfo{person}{Kang Zhao}, \bibinfo{person}{Keyu Yan}, \bibinfo{person}{Lianghua Huang}, \bibinfo{person}{Mengyang Feng}, \bibinfo{person}{Ningyi Zhang}, \bibinfo{person}{Pandeng Li}, \bibinfo{person}{Pingyu Wu}, \bibinfo{person}{Ruihang Chu}, \bibinfo{person}{Ruili Feng}, \bibinfo{person}{Shiwei Zhang}, \bibinfo{person}{Siyang Sun}, \bibinfo{person}{Tao Fang}, \bibinfo{person}{Tianxing Wang}, \bibinfo{person}{Tianyi Gui}, \bibinfo{person}{Tingyu Weng}, \bibinfo{person}{Tong Shen}, \bibinfo{person}{Wei Lin},
  \bibinfo{person}{Wei Wang}, \bibinfo{person}{Wei Wang}, \bibinfo{person}{Wenmeng Zhou}, \bibinfo{person}{Wente Wang}, \bibinfo{person}{Wenting Shen}, \bibinfo{person}{Wenyuan Yu}, \bibinfo{person}{Xianzhong Shi}, \bibinfo{person}{Xiaoming Huang}, \bibinfo{person}{Xin Xu}, \bibinfo{person}{Yan Kou}, \bibinfo{person}{Yangyu Lv}, \bibinfo{person}{Yifei Li}, \bibinfo{person}{Yijing Liu}, \bibinfo{person}{Yiming Wang}, \bibinfo{person}{Yingya Zhang}, \bibinfo{person}{Yitong Huang}, \bibinfo{person}{Yong Li}, \bibinfo{person}{You Wu}, \bibinfo{person}{Yu Liu}, \bibinfo{person}{Yulin Pan}, \bibinfo{person}{Yun Zheng}, \bibinfo{person}{Yuntao Hong}, \bibinfo{person}{Yupeng Shi}, \bibinfo{person}{Yutong Feng}, \bibinfo{person}{Zeyinzi Jiang}, \bibinfo{person}{Zhen Han}, \bibinfo{person}{Zhi-Fan Wu}, {and} \bibinfo{person}{Ziyu Liu}.} \bibinfo{year}{2025}\natexlab{a}.
\newblock \showarticletitle{Wan: Open and Advanced Large-Scale Video Generative Models}.
\newblock \bibinfo{journal}{\emph{arXiv preprint arXiv:2503.20314}} (\bibinfo{year}{2025}).
\newblock


\bibitem[Wang et~al\mbox{.}(2025b)]%
        {wang2025videoscene}
\bibfield{author}{\bibinfo{person}{Hanyang Wang}, \bibinfo{person}{Fangfu Liu}, \bibinfo{person}{Jiawei Chi}, {and} \bibinfo{person}{Yueqi Duan}.} \bibinfo{year}{2025}\natexlab{b}.
\newblock \showarticletitle{VideoScene: Distilling Video Diffusion Model to Generate 3D Scenes in One Step}.
\newblock \bibinfo{journal}{\emph{arXiv preprint arXiv:2504.01956}} (\bibinfo{year}{2025}).
\newblock


\bibitem[Wang et~al\mbox{.}(2021)]%
        {wang2021neus}
\bibfield{author}{\bibinfo{person}{Peng Wang}, \bibinfo{person}{Lingjie Liu}, \bibinfo{person}{Yuan Liu}, \bibinfo{person}{Christian Theobalt}, \bibinfo{person}{Taku Komura}, {and} \bibinfo{person}{Wenping Wang}.} \bibinfo{year}{2021}\natexlab{}.
\newblock \showarticletitle{NeuS: Learning Neural Implicit Surfaces by Volume Rendering for Multi-view Reconstruction}.
\newblock \bibinfo{journal}{\emph{arXiv preprint arXiv:2106.10689}} (\bibinfo{year}{2021}).
\newblock


\bibitem[Wang and Shi(2023)]%
        {wang2023imagedream}
\bibfield{author}{\bibinfo{person}{Peng Wang} {and} \bibinfo{person}{Yichun Shi}.} \bibinfo{year}{2023}\natexlab{}.
\newblock \showarticletitle{Imagedream: Image-prompt multi-view diffusion for 3d generation}.
\newblock \bibinfo{journal}{\emph{arXiv preprint arXiv:2312.02201}} (\bibinfo{year}{2023}).
\newblock


\bibitem[Wang et~al\mbox{.}(2024a)]%
        {WangCVPR2024}
\bibfield{author}{\bibinfo{person}{Shaofei Wang}, \bibinfo{person}{Bo\v{z}idar Anti\'{c}}, \bibinfo{person}{Andreas Geiger}, {and} \bibinfo{person}{Siyu Tang}.} \bibinfo{year}{2024}\natexlab{a}.
\newblock \showarticletitle{IntrinsicAvatar: Physically Based Inverse Rendering of Dynamic Humans from Monocular Videos via Explicit Ray Tracing}. In \bibinfo{booktitle}{\emph{Proceedings IEEE Conf. on Computer Vision and Pattern Recognition (CVPR)}}.
\newblock


\bibitem[Wang et~al\mbox{.}(2022)]%
        {wang2022arah}
\bibfield{author}{\bibinfo{person}{Shaofei Wang}, \bibinfo{person}{Katja Schwarz}, \bibinfo{person}{Andreas Geiger}, {and} \bibinfo{person}{Siyu Tang}.} \bibinfo{year}{2022}\natexlab{}.
\newblock \showarticletitle{Arah: Animatable volume rendering of articulated human sdfs}. In \bibinfo{booktitle}{\emph{European conference on computer vision}}. Springer, \bibinfo{pages}{1--19}.
\newblock


\bibitem[Wang et~al\mbox{.}(2025c)]%
        {wang2025freetimegs}
\bibfield{author}{\bibinfo{person}{Yifan Wang}, \bibinfo{person}{Peishan Yang}, \bibinfo{person}{Zhen Xu}, \bibinfo{person}{Jiaming Sun}, \bibinfo{person}{Zhanhua Zhang}, \bibinfo{person}{Yong Chen}, \bibinfo{person}{Hujun Bao}, \bibinfo{person}{Sida Peng}, {and} \bibinfo{person}{Xiaowei Zhou}.} \bibinfo{year}{2025}\natexlab{c}.
\newblock \showarticletitle{FreeTimeGS: Free Gaussian Primitives at Anytime Anywhere for Dynamic Scene Reconstruction}. In \bibinfo{booktitle}{\emph{CVPR}}.
\newblock
\urldef\tempurl%
\url{https://zju3dv.github.io/freetimegs}
\showURL{%
\tempurl}


\bibitem[Wang et~al\mbox{.}(2004)]%
        {ssim}
\bibfield{author}{\bibinfo{person}{Zhou Wang}, \bibinfo{person}{Alan~C Bovik}, \bibinfo{person}{Hamid~R Sheikh}, {and} \bibinfo{person}{Eero~P Simoncelli}.} \bibinfo{year}{2004}\natexlab{}.
\newblock \showarticletitle{Image quality assessment: from error visibility to structural similarity}.
\newblock \bibinfo{journal}{\emph{IEEE transactions on image processing}} \bibinfo{volume}{13}, \bibinfo{number}{4} (\bibinfo{year}{2004}), \bibinfo{pages}{600--612}.
\newblock


\bibitem[Wang et~al\mbox{.}(2024b)]%
        {wang2024motionctrl}
\bibfield{author}{\bibinfo{person}{Zhouxia Wang}, \bibinfo{person}{Ziyang Yuan}, \bibinfo{person}{Xintao Wang}, \bibinfo{person}{Yaowei Li}, \bibinfo{person}{Tianshui Chen}, \bibinfo{person}{Menghan Xia}, \bibinfo{person}{Ping Luo}, {and} \bibinfo{person}{Ying Shan}.} \bibinfo{year}{2024}\natexlab{b}.
\newblock \showarticletitle{Motionctrl: A unified and flexible motion controller for video generation}. In \bibinfo{booktitle}{\emph{ACM SIGGRAPH 2024 Conference Papers}}. \bibinfo{pages}{1--11}.
\newblock


\bibitem[Watson et~al\mbox{.}(2022)]%
        {watson2022novel}
\bibfield{author}{\bibinfo{person}{Daniel Watson}, \bibinfo{person}{William Chan}, \bibinfo{person}{Ricardo Martin-Brualla}, \bibinfo{person}{Jonathan Ho}, \bibinfo{person}{Andrea Tagliasacchi}, {and} \bibinfo{person}{Mohammad Norouzi}.} \bibinfo{year}{2022}\natexlab{}.
\newblock \showarticletitle{Novel view synthesis with diffusion models}.
\newblock \bibinfo{journal}{\emph{arXiv preprint arXiv:2210.04628}} (\bibinfo{year}{2022}).
\newblock


\bibitem[Wen et~al\mbox{.}(2024)]%
        {wen2024gomavatar}
\bibfield{author}{\bibinfo{person}{Jing Wen}, \bibinfo{person}{Xiaoming Zhao}, \bibinfo{person}{Zhongzheng Ren}, \bibinfo{person}{Alexander~G Schwing}, {and} \bibinfo{person}{Shenlong Wang}.} \bibinfo{year}{2024}\natexlab{}.
\newblock \showarticletitle{Gomavatar: Efficient animatable human modeling from monocular video using gaussians-on-mesh}. In \bibinfo{booktitle}{\emph{Proceedings of the IEEE/CVF Conference on Computer Vision and Pattern Recognition}}. \bibinfo{pages}{2059--2069}.
\newblock


\bibitem[Weng et~al\mbox{.}(2022)]%
        {weng2022humannerf}
\bibfield{author}{\bibinfo{person}{Chung-Yi Weng}, \bibinfo{person}{Brian Curless}, \bibinfo{person}{Pratul~P Srinivasan}, \bibinfo{person}{Jonathan~T Barron}, {and} \bibinfo{person}{Ira Kemelmacher-Shlizerman}.} \bibinfo{year}{2022}\natexlab{}.
\newblock \showarticletitle{Humannerf: Free-viewpoint rendering of moving people from monocular video}. In \bibinfo{booktitle}{\emph{Proceedings of the IEEE/CVF conference on computer vision and pattern Recognition}}. \bibinfo{pages}{16210--16220}.
\newblock


\bibitem[Wu et~al\mbox{.}(2024b)]%
        {wu20244d}
\bibfield{author}{\bibinfo{person}{Guanjun Wu}, \bibinfo{person}{Taoran Yi}, \bibinfo{person}{Jiemin Fang}, \bibinfo{person}{Lingxi Xie}, \bibinfo{person}{Xiaopeng Zhang}, \bibinfo{person}{Wei Wei}, \bibinfo{person}{Wenyu Liu}, \bibinfo{person}{Qi Tian}, {and} \bibinfo{person}{Xinggang Wang}.} \bibinfo{year}{2024}\natexlab{b}.
\newblock \showarticletitle{4d gaussian splatting for real-time dynamic scene rendering}. In \bibinfo{booktitle}{\emph{Proceedings of the IEEE/CVF conference on computer vision and pattern recognition}}. \bibinfo{pages}{20310--20320}.
\newblock


\bibitem[Wu et~al\mbox{.}(2020)]%
        {wu2020multi}
\bibfield{author}{\bibinfo{person}{Minye Wu}, \bibinfo{person}{Yuehao Wang}, \bibinfo{person}{Qiang Hu}, {and} \bibinfo{person}{Jingyi Yu}.} \bibinfo{year}{2020}\natexlab{}.
\newblock \showarticletitle{Multi-view neural human rendering}. In \bibinfo{booktitle}{\emph{Proceedings of the IEEE/CVF Conference on Computer Vision and Pattern Recognition}}. \bibinfo{pages}{1682--1691}.
\newblock


\bibitem[Wu et~al\mbox{.}(2024a)]%
        {wu2024cat4d}
\bibfield{author}{\bibinfo{person}{Rundi Wu}, \bibinfo{person}{Ruiqi Gao}, \bibinfo{person}{Ben Poole}, \bibinfo{person}{Alex Trevithick}, \bibinfo{person}{Changxi Zheng}, \bibinfo{person}{Jonathan~T Barron}, {and} \bibinfo{person}{Aleksander Holynski}.} \bibinfo{year}{2024}\natexlab{a}.
\newblock \showarticletitle{Cat4d: Create anything in 4d with multi-view video diffusion models}.
\newblock \bibinfo{journal}{\emph{arXiv preprint arXiv:2411.18613}} (\bibinfo{year}{2024}).
\newblock


\bibitem[Wu et~al\mbox{.}(2024c)]%
        {wu2024sc4d}
\bibfield{author}{\bibinfo{person}{Zijie Wu}, \bibinfo{person}{Chaohui Yu}, \bibinfo{person}{Yanqin Jiang}, \bibinfo{person}{Chenjie Cao}, \bibinfo{person}{Fan Wang}, {and} \bibinfo{person}{Xiang Bai}.} \bibinfo{year}{2024}\natexlab{c}.
\newblock \showarticletitle{Sc4d: Sparse-controlled video-to-4d generation and motion transfer}. In \bibinfo{booktitle}{\emph{European Conference on Computer Vision}}. Springer, \bibinfo{pages}{361--379}.
\newblock


\bibitem[Xiang et~al\mbox{.}(2023)]%
        {xiang20233d}
\bibfield{author}{\bibinfo{person}{Jianfeng Xiang}, \bibinfo{person}{Jiaolong Yang}, \bibinfo{person}{Binbin Huang}, {and} \bibinfo{person}{Xin Tong}.} \bibinfo{year}{2023}\natexlab{}.
\newblock \showarticletitle{3d-aware image generation using 2d diffusion models}. In \bibinfo{booktitle}{\emph{Proceedings of the IEEE/CVF International Conference on Computer Vision}}. \bibinfo{pages}{2383--2393}.
\newblock


\bibitem[Xie et~al\mbox{.}(2024)]%
        {xie2024sv4d}
\bibfield{author}{\bibinfo{person}{Yiming Xie}, \bibinfo{person}{Chun-Han Yao}, \bibinfo{person}{Vikram Voleti}, \bibinfo{person}{Huaizu Jiang}, {and} \bibinfo{person}{Varun Jampani}.} \bibinfo{year}{2024}\natexlab{}.
\newblock \showarticletitle{Sv4d: Dynamic 3d content generation with multi-frame and multi-view consistency}.
\newblock \bibinfo{journal}{\emph{arXiv preprint arXiv:2407.17470}} (\bibinfo{year}{2024}).
\newblock


\bibitem[Xing et~al\mbox{.}(2023)]%
        {xing2023dynamicrafter}
\bibfield{author}{\bibinfo{person}{Jinbo Xing}, \bibinfo{person}{Menghan Xia}, \bibinfo{person}{Yong Zhang}, \bibinfo{person}{Haoxin Chen}, \bibinfo{person}{Xintao Wang}, \bibinfo{person}{Tien-Tsin Wong}, {and} \bibinfo{person}{Ying Shan}.} \bibinfo{year}{2023}\natexlab{}.
\newblock \showarticletitle{DynamiCrafter: Animating Open-domain Images with Video Diffusion Priors}.
\newblock  (\bibinfo{year}{2023}).
\newblock
\showeprint[arxiv]{2310.12190}~[cs.CV]


\bibitem[Xiong et~al\mbox{.}(2024)]%
        {xiong2024mvhumannet}
\bibfield{author}{\bibinfo{person}{Zhangyang Xiong}, \bibinfo{person}{Chenghong Li}, \bibinfo{person}{Kenkun Liu}, \bibinfo{person}{Hongjie Liao}, \bibinfo{person}{Jianqiao Hu}, \bibinfo{person}{Junyi Zhu}, \bibinfo{person}{Shuliang Ning}, \bibinfo{person}{Lingteng Qiu}, \bibinfo{person}{Chongjie Wang}, \bibinfo{person}{Shijie Wang}, {et~al\mbox{.}}} \bibinfo{year}{2024}\natexlab{}.
\newblock \showarticletitle{MVHumanNet: A Large-scale Dataset of Multi-view Daily Dressing Human Captures}. In \bibinfo{booktitle}{\emph{Proceedings of the IEEE/CVF Conference on Computer Vision and Pattern Recognition}}.
\newblock


\bibitem[Xu et~al\mbox{.}(2022)]%
        {xu2022point}
\bibfield{author}{\bibinfo{person}{Qiangeng Xu}, \bibinfo{person}{Zexiang Xu}, \bibinfo{person}{Julien Philip}, \bibinfo{person}{Sai Bi}, \bibinfo{person}{Zhixin Shu}, \bibinfo{person}{Kalyan Sunkavalli}, {and} \bibinfo{person}{Ulrich Neumann}.} \bibinfo{year}{2022}\natexlab{}.
\newblock \showarticletitle{Point-nerf: Point-based neural radiance fields}. In \bibinfo{booktitle}{\emph{Proceedings of the IEEE/CVF conference on computer vision and pattern recognition}}. \bibinfo{pages}{5438--5448}.
\newblock


\bibitem[Xu et~al\mbox{.}(2024a)]%
        {xu20244k4d}
\bibfield{author}{\bibinfo{person}{Zhen Xu}, \bibinfo{person}{Sida Peng}, \bibinfo{person}{Haotong Lin}, \bibinfo{person}{Guangzhao He}, \bibinfo{person}{Jiaming Sun}, \bibinfo{person}{Yujun Shen}, \bibinfo{person}{Hujun Bao}, {and} \bibinfo{person}{Xiaowei Zhou}.} \bibinfo{year}{2024}\natexlab{a}.
\newblock \showarticletitle{4k4d: Real-time 4d view synthesis at 4k resolution}. In \bibinfo{booktitle}{\emph{Proceedings of the IEEE/CVF conference on computer vision and pattern recognition}}. \bibinfo{pages}{20029--20040}.
\newblock


\bibitem[Xu et~al\mbox{.}(2024b)]%
        {xu2024longvolcap}
\bibfield{author}{\bibinfo{person}{Zhen Xu}, \bibinfo{person}{Yinghao Xu}, \bibinfo{person}{Zhiyuan Yu}, \bibinfo{person}{Sida Peng}, \bibinfo{person}{Jiaming Sun}, \bibinfo{person}{Hujun Bao}, {and} \bibinfo{person}{Xiaowei Zhou}.} \bibinfo{year}{2024}\natexlab{b}.
\newblock \showarticletitle{Representing Long Volumetric Video with Temporal Gaussian Hierarchy}.
\newblock \bibinfo{journal}{\emph{ACM Transactions on Graphics}} \bibinfo{volume}{43}, \bibinfo{number}{6} (\bibinfo{date}{November} \bibinfo{year}{2024}).
\newblock
\urldef\tempurl%
\url{https://zju3dv.github.io/longvolcap}
\showURL{%
\tempurl}


\bibitem[Yang et~al\mbox{.}(2024a)]%
        {yang2024cogvideox}
\bibfield{author}{\bibinfo{person}{Zhuoyi Yang}, \bibinfo{person}{Jiayan Teng}, \bibinfo{person}{Wendi Zheng}, \bibinfo{person}{Ming Ding}, \bibinfo{person}{Shiyu Huang}, \bibinfo{person}{Jiazheng Xu}, \bibinfo{person}{Yuanming Yang}, \bibinfo{person}{Wenyi Hong}, \bibinfo{person}{Xiaohan Zhang}, \bibinfo{person}{Guanyu Feng}, {et~al\mbox{.}}} \bibinfo{year}{2024}\natexlab{a}.
\newblock \showarticletitle{Cogvideox: Text-to-video diffusion models with an expert transformer}.
\newblock \bibinfo{journal}{\emph{arXiv preprint arXiv:2408.06072}} (\bibinfo{year}{2024}).
\newblock


\bibitem[Yang et~al\mbox{.}(2024b)]%
        {yang2023gs4d}
\bibfield{author}{\bibinfo{person}{Zeyu Yang}, \bibinfo{person}{Hongye Yang}, \bibinfo{person}{Zijie Pan}, {and} \bibinfo{person}{Li Zhang}.} \bibinfo{year}{2024}\natexlab{b}.
\newblock \showarticletitle{Real-time Photorealistic Dynamic Scene Representation and Rendering with 4D Gaussian Splatting}. In \bibinfo{booktitle}{\emph{International Conference on Learning Representations (ICLR)}}.
\newblock


\bibitem[Yariv et~al\mbox{.}(2021)]%
        {yariv2021volume}
\bibfield{author}{\bibinfo{person}{Lior Yariv}, \bibinfo{person}{Jiatao Gu}, \bibinfo{person}{Yoni Kasten}, {and} \bibinfo{person}{Yaron Lipman}.} \bibinfo{year}{2021}\natexlab{}.
\newblock \showarticletitle{Volume rendering of neural implicit surfaces}.
\newblock \bibinfo{journal}{\emph{Advances in Neural Information Processing Systems}}  \bibinfo{volume}{34} (\bibinfo{year}{2021}), \bibinfo{pages}{4805--4815}.
\newblock


\bibitem[YU et~al\mbox{.}(2025)]%
        {yu2025trajectorycrafter}
\bibfield{author}{\bibinfo{person}{Mark YU}, \bibinfo{person}{Wenbo Hu}, \bibinfo{person}{Jinbo Xing}, {and} \bibinfo{person}{Ying Shan}.} \bibinfo{year}{2025}\natexlab{}.
\newblock \showarticletitle{TrajectoryCrafter: Redirecting Camera Trajectory for Monocular Videos via Diffusion Models}.
\newblock \bibinfo{journal}{\emph{arXiv preprint arXiv:2503.05638}} (\bibinfo{year}{2025}).
\newblock


\bibitem[Yu et~al\mbox{.}(2021)]%
        {yu2021function4d}
\bibfield{author}{\bibinfo{person}{Tao Yu}, \bibinfo{person}{Zerong Zheng}, \bibinfo{person}{Kaiwen Guo}, \bibinfo{person}{Pengpeng Liu}, \bibinfo{person}{Qionghai Dai}, {and} \bibinfo{person}{Yebin Liu}.} \bibinfo{year}{2021}\natexlab{}.
\newblock \showarticletitle{Function4d: Real-time human volumetric capture from very sparse consumer rgbd sensors}. In \bibinfo{booktitle}{\emph{Proceedings of the IEEE/CVF conference on computer vision and pattern recognition}}. \bibinfo{pages}{5746--5756}.
\newblock


\bibitem[Yu et~al\mbox{.}(2024)]%
        {yu2024viewcrafter}
\bibfield{author}{\bibinfo{person}{Wangbo Yu}, \bibinfo{person}{Jinbo Xing}, \bibinfo{person}{Li Yuan}, \bibinfo{person}{Wenbo Hu}, \bibinfo{person}{Xiaoyu Li}, \bibinfo{person}{Zhipeng Huang}, \bibinfo{person}{Xiangjun Gao}, \bibinfo{person}{Tien-Tsin Wong}, \bibinfo{person}{Ying Shan}, {and} \bibinfo{person}{Yonghong Tian}.} \bibinfo{year}{2024}\natexlab{}.
\newblock \showarticletitle{ViewCrafter: Taming Video Diffusion Models for High-fidelity Novel View Synthesis}.
\newblock \bibinfo{journal}{\emph{arXiv preprint arXiv:2409.02048}} (\bibinfo{year}{2024}).
\newblock


\bibitem[Yu et~al\mbox{.}(2022)]%
        {yu2022monosdf}
\bibfield{author}{\bibinfo{person}{Zehao Yu}, \bibinfo{person}{Songyou Peng}, \bibinfo{person}{Michael Niemeyer}, \bibinfo{person}{Torsten Sattler}, {and} \bibinfo{person}{Andreas Geiger}.} \bibinfo{year}{2022}\natexlab{}.
\newblock \showarticletitle{Monosdf: Exploring monocular geometric cues for neural implicit surface reconstruction}.
\newblock \bibinfo{journal}{\emph{Advances in neural information processing systems}}  \bibinfo{volume}{35} (\bibinfo{year}{2022}), \bibinfo{pages}{25018--25032}.
\newblock


\bibitem[Zeng et~al\mbox{.}(2024)]%
        {zeng2024stag4d}
\bibfield{author}{\bibinfo{person}{Yifei Zeng}, \bibinfo{person}{Yanqin Jiang}, \bibinfo{person}{Siyu Zhu}, \bibinfo{person}{Yuanxun Lu}, \bibinfo{person}{Youtian Lin}, \bibinfo{person}{Hao Zhu}, \bibinfo{person}{Weiming Hu}, \bibinfo{person}{Xun Cao}, {and} \bibinfo{person}{Yao Yao}.} \bibinfo{year}{2024}\natexlab{}.
\newblock \showarticletitle{Stag4d: Spatial-temporal anchored generative 4d gaussians}. In \bibinfo{booktitle}{\emph{European Conference on Computer Vision}}. Springer, \bibinfo{pages}{163--179}.
\newblock


\bibitem[Zhang et~al\mbox{.}(2018)]%
        {lpips}
\bibfield{author}{\bibinfo{person}{Richard Zhang}, \bibinfo{person}{Phillip Isola}, \bibinfo{person}{Alexei~A Efros}, \bibinfo{person}{Eli Shechtman}, {and} \bibinfo{person}{Oliver Wang}.} \bibinfo{year}{2018}\natexlab{}.
\newblock \showarticletitle{The unreasonable effectiveness of deep features as a perceptual metric}. In \bibinfo{booktitle}{\emph{Proceedings of the IEEE conference on computer vision and pattern recognition}}. \bibinfo{pages}{586--595}.
\newblock


\bibitem[Zhao et~al\mbox{.}(2022a)]%
        {zhao2022human}
\bibfield{author}{\bibinfo{person}{Fuqiang Zhao}, \bibinfo{person}{Yuheng Jiang}, \bibinfo{person}{Kaixin Yao}, \bibinfo{person}{Jiakai Zhang}, \bibinfo{person}{Liao Wang}, \bibinfo{person}{Haizhao Dai}, \bibinfo{person}{Yuhui Zhong}, \bibinfo{person}{Yingliang Zhang}, \bibinfo{person}{Minye Wu}, \bibinfo{person}{Lan Xu}, {et~al\mbox{.}}} \bibinfo{year}{2022}\natexlab{a}.
\newblock \showarticletitle{Human performance modeling and rendering via neural animated mesh}.
\newblock \bibinfo{journal}{\emph{ACM Transactions on Graphics (TOG)}} \bibinfo{volume}{41}, \bibinfo{number}{6} (\bibinfo{year}{2022}), \bibinfo{pages}{1--17}.
\newblock


\bibitem[Zhao et~al\mbox{.}(2022b)]%
        {zhao2022humannerf}
\bibfield{author}{\bibinfo{person}{Fuqiang Zhao}, \bibinfo{person}{Wei Yang}, \bibinfo{person}{Jiakai Zhang}, \bibinfo{person}{Pei Lin}, \bibinfo{person}{Yingliang Zhang}, \bibinfo{person}{Jingyi Yu}, {and} \bibinfo{person}{Lan Xu}.} \bibinfo{year}{2022}\natexlab{b}.
\newblock \showarticletitle{Humannerf: Efficiently generated human radiance field from sparse inputs}. In \bibinfo{booktitle}{\emph{Proceedings of the IEEE/CVF Conference on Computer Vision and Pattern Recognition}}. \bibinfo{pages}{7743--7753}.
\newblock


\bibitem[Zhao et~al\mbox{.}(2025)]%
        {zhao2025surfel}
\bibfield{author}{\bibinfo{person}{Yiqun Zhao}, \bibinfo{person}{Chenming Wu}, \bibinfo{person}{Binbin Huang}, \bibinfo{person}{Yihao Zhi}, \bibinfo{person}{Chen Zhao}, \bibinfo{person}{Jingdong Wang}, {and} \bibinfo{person}{Shenghua Gao}.} \bibinfo{year}{2025}\natexlab{}.
\newblock \showarticletitle{Surfel-based Gaussian Inverse Rendering for Fast and Relightable Dynamic Human Reconstruction from Monocular Videos}.
\newblock \bibinfo{journal}{\emph{IEEE Transactions on Pattern Analysis and Machine Intelligence}} (\bibinfo{year}{2025}).
\newblock


\bibitem[Zheng et~al\mbox{.}(2024)]%
        {zheng2024gps}
\bibfield{author}{\bibinfo{person}{Shunyuan Zheng}, \bibinfo{person}{Boyao Zhou}, \bibinfo{person}{Ruizhi Shao}, \bibinfo{person}{Boning Liu}, \bibinfo{person}{Shengping Zhang}, \bibinfo{person}{Liqiang Nie}, {and} \bibinfo{person}{Yebin Liu}.} \bibinfo{year}{2024}\natexlab{}.
\newblock \showarticletitle{Gps-gaussian: Generalizable pixel-wise 3d gaussian splatting for real-time human novel view synthesis}. In \bibinfo{booktitle}{\emph{Proceedings of the IEEE/CVF conference on computer vision and pattern recognition}}. \bibinfo{pages}{19680--19690}.
\newblock


\bibitem[Zheng et~al\mbox{.}(2022)]%
        {zheng2022structured}
\bibfield{author}{\bibinfo{person}{Zerong Zheng}, \bibinfo{person}{Han Huang}, \bibinfo{person}{Tao Yu}, \bibinfo{person}{Hongwen Zhang}, \bibinfo{person}{Yandong Guo}, {and} \bibinfo{person}{Yebin Liu}.} \bibinfo{year}{2022}\natexlab{}.
\newblock \showarticletitle{Structured local radiance fields for human avatar modeling}. In \bibinfo{booktitle}{\emph{Proceedings of the IEEE/CVF Conference on Computer Vision and Pattern Recognition}}. \bibinfo{pages}{15893--15903}.
\newblock


\bibitem[Zheng et~al\mbox{.}(2023)]%
        {zheng2023avatarrex}
\bibfield{author}{\bibinfo{person}{Zerong Zheng}, \bibinfo{person}{Xiaochen Zhao}, \bibinfo{person}{Hongwen Zhang}, \bibinfo{person}{Boning Liu}, {and} \bibinfo{person}{Yebin Liu}.} \bibinfo{year}{2023}\natexlab{}.
\newblock \showarticletitle{Avatarrex: Real-time expressive full-body avatars}.
\newblock \bibinfo{journal}{\emph{ACM Transactions on Graphics (TOG)}} \bibinfo{volume}{42}, \bibinfo{number}{4} (\bibinfo{year}{2023}), \bibinfo{pages}{1--19}.
\newblock


\bibitem[Zhi et~al\mbox{.}(2022)]%
        {zhi2022dual}
\bibfield{author}{\bibinfo{person}{Yihao Zhi}, \bibinfo{person}{Shenhan Qian}, \bibinfo{person}{Xinhao Yan}, {and} \bibinfo{person}{Shenghua Gao}.} \bibinfo{year}{2022}\natexlab{}.
\newblock \showarticletitle{Dual-space nerf: Learning animatable avatars and scene lighting in separate spaces}. In \bibinfo{booktitle}{\emph{2022 International Conference on 3D Vision (3DV)}}. IEEE, \bibinfo{pages}{1--10}.
\newblock


\bibitem[Zhi et~al\mbox{.}(2025)]%
        {zhi2025strugauavatar}
\bibfield{author}{\bibinfo{person}{Yihao Zhi}, \bibinfo{person}{Wanhu Sun}, \bibinfo{person}{Jiahao Chang}, \bibinfo{person}{Chongjie Ye}, \bibinfo{person}{Wensen Feng}, {and} \bibinfo{person}{Xiaoguang Han}.} \bibinfo{year}{2025}\natexlab{}.
\newblock \showarticletitle{StruGauAvatar: Learning Structured 3D Gaussians for Animatable Avatars from Monocular Videos}.
\newblock \bibinfo{journal}{\emph{IEEE Transactions on Visualization and Computer Graphics}} (\bibinfo{year}{2025}).
\newblock


\bibitem[Zhu et~al\mbox{.}(2024)]%
        {zhu2024champ}
\bibfield{author}{\bibinfo{person}{Shenhao Zhu}, \bibinfo{person}{Junming~Leo Chen}, \bibinfo{person}{Zuozhuo Dai}, \bibinfo{person}{Yinghui Xu}, \bibinfo{person}{Xun Cao}, \bibinfo{person}{Yao Yao}, \bibinfo{person}{Hao Zhu}, {and} \bibinfo{person}{Siyu Zhu}.} \bibinfo{year}{2024}\natexlab{}.
\newblock \bibinfo{title}{Champ: Controllable and Consistent Human Image Animation with 3D Parametric Guidance}.
\newblock
\showeprint[arxiv]{2403.14781}~[cs.CV]


\bibitem[Zhuang et~al\mbox{.}(2024)]%
        {zhuang2024idol}
\bibfield{author}{\bibinfo{person}{Yiyu Zhuang}, \bibinfo{person}{Jiaxi Lv}, \bibinfo{person}{Hao Wen}, \bibinfo{person}{Qing Shuai}, \bibinfo{person}{Ailing Zeng}, \bibinfo{person}{Hao Zhu}, \bibinfo{person}{Shifeng Chen}, \bibinfo{person}{Yujiu Yang}, \bibinfo{person}{Xun Cao}, {and} \bibinfo{person}{Wei Liu}.} \bibinfo{year}{2024}\natexlab{}.
\newblock \showarticletitle{IDOL: Instant Photorealistic 3D Human Creation from a Single Image}.
\newblock \bibinfo{journal}{\emph{arXiv preprint arXiv:2412.14963}} (\bibinfo{year}{2024}).
\newblock


\end{thebibliography}

\end{document}